\documentclass{article}
\usepackage{style/unites}

\usepackage[utf8]{inputenc}
\usepackage[T1]{fontenc}
\usepackage{microtype}

\usepackage{amsmath}
\usepackage{amssymb}
\usepackage{amsfonts}
\usepackage{amsthm}
\usepackage{mathtools}
\usepackage{mathrsfs}
\usepackage{physics}
\usepackage{braket}
\usepackage{slashed}
\usepackage{nicefrac}
\usepackage{textcomp}
\usepackage{dsfont}
\usepackage{bbm}
\usepackage{bm}

\usepackage{graphicx}
\usepackage{subcaption}
\usepackage[export]{adjustbox}
\usepackage{float}
\usepackage{booktabs}
\usepackage{dcolumn}
\newcolumntype{d}[1]{D{.}{.}{#1}}
\usepackage{bigstrut, tabularx, multirow, makecell, diagbox}
\usepackage{colortbl}
\usepackage{tabularray}
\UseTblrLibrary{booktabs}
\usepackage{threeparttable}
\usepackage{tablefootnote}
\usepackage{fontawesome5}

\usepackage{placeins}
\usepackage{caption}
\usepackage{footnote}
\usepackage{enumitem}
\usepackage{multicol}
\usepackage{xspace}
\usepackage{titletoc}
\usepackage{titlesec}
\usepackage[bottom]{footmisc}
\usepackage{setspace}

\usepackage{wrapfig}
\usepackage{tikz}
\usepackage{quantikz}
\usepackage{dashbox}
\usepackage{mdframed}
\usepackage{marvosym}
\usepackage{pifont}
\usepackage{CJK}
\usepackage{url}

\usepackage{algorithm}       
\usepackage{algpseudocode}   

\usepackage[table,x11names]{xcolor}
\usepackage[most]{tcolorbox}
\tcbuselibrary{breakable}
\usetikzlibrary{decorations.pathreplacing, fit}

\definecolor{primaryblue}{HTML}{0066CC}
\definecolor{accentcyan}{HTML}{00D4AA}
\definecolor{warmorange}{HTML}{FF6B35}
\definecolor{deepgray}{HTML}{2C3E50}
\definecolor{lightgray}{HTML}{F8F9FA}
\definecolor{gradientstart}{HTML}{667eea}
\definecolor{gradientend}{HTML}{764ba2}

\definecolor{citecolor}{HTML}{0071bc}
\definecolor{citeblue}{RGB}{0, 113, 188}
\definecolor{linkcolor}{HTML}{9A4D92}
\definecolor{firebrick}{rgb}{0.698,0.133,0.133}

\definecolor{paleviolet}{HTML}{E1EEFC}
\definecolor{CarolinaUltraLight}{HTML}{E7F4FC}
\definecolor{lightgrey}{RGB}{247, 247, 247}
\definecolor{shadecolor}{HTML}{EFEFEF}
\definecolor{lightyellow}{rgb}{1.0, 0.95, 0.7}
\definecolor{lightblue}{rgb}{0.90, 0.95, 1.0}
\definecolor{light-gray}{gray}{0.95}

\definecolor{darkgrey}{rgb}{0.5, 0.5, 0.5}
\definecolor{darkgreen}{rgb}{0, 0.5, 0}
\definecolor{mydarkblue}{rgb}{0,0.08,0.45}
\definecolor{mydarkblue2}{rgb}{0.133, 0.133, 0.698}
\definecolor{echodrk}{HTML}{0099cc}
\definecolor{mymauve}{rgb}{0.58,0,0.82}
\definecolor{midnightblue}{rgb}{0.1,0.1,0.44}
\definecolor{oxfordblue}{rgb}{0.0,0.13,0.28}
\definecolor{prussianblue}{rgb}{0.0,0.19,0.33}
\definecolor{coolteal}{rgb}{0, 0.45, 0.45}
\definecolor{olive}{rgb}{0.1, 0.3, 0}
\definecolor{mypurple}{rgb}{0.5,0,0.5}
\definecolor{almond}{rgb}{0.94, 0.87, 0.8}

\definecolor{blue_ampEncoding}{HTML}{DAE8FC}
\definecolor{green_encoder}{HTML}{D5E8D4}
\definecolor{purple_decoder}{HTML}{E1D5E7}
\definecolor{yellow_measure}{HTML}{FFF2CC}
\definecolor{gray_block}{HTML}{F5F5F5}
\definecolor{pink_dru}{HTML}{FAD9D5}
\definecolor{orange_v}{HTML}{FAD7AC}

\definecolor{colorA}{rgb}{1,0,0}
\definecolor{colorB}{rgb}{0,0.3,1}
\definecolor{colorC}{rgb}{0.9,0.8,0.2}
\definecolor{colorD}{rgb}{0,0.65,0}
\definecolor{lesslightgray}{rgb}{0.5,0.5,0.5}
\definecolor{fundamental}{RGB}{55, 110, 111}
\definecolor{Gred}{RGB}{219, 50, 54}
\definecolor{ToCgreen}{RGB}{0, 128, 0}
\definecolor{Sepia}{RGB}{112, 66, 20}
\definecolor{Dblue}{rgb}{0,0.08,0.45}
\definecolor{Blue}{rgb}{0, 0, 0.8}
\definecolor{blue}{rgb}{0,0,1}
\definecolor{UNCblue!10}{rgb}{0.84,0.91,0.98}
\definecolor{RowAlt}{rgb}{0.98,0.98,0.99}

\definecolor{CarolinaBlue}{HTML}{7BAFD4}        
\definecolor{CarolinaLightBlue}{HTML}{B3D4E5}   
\definecolor{CarolinaUltraLight}{HTML}{E8F4F8}  
\definecolor{CarolinaText}{HTML}{1C2B33}        

\usepackage[pagebackref=true,breaklinks=true,colorlinks,hyperfootnotes=false]{hyperref}
\hypersetup{
  colorlinks,
  citecolor=citeblue,
  linkcolor=citeblue,
  urlcolor=citeblue
}
\usepackage[nameinlink,capitalize,noabbrev]{cleveref}

\titlespacing\section{0pt}{4pt plus 4pt minus 2pt}{-2pt plus 2pt minus 2pt}
\titlespacing\subsection{0pt}{2pt plus 4pt minus 2pt}{-2pt plus 2pt minus 2pt}
\titlespacing\subsubsection{0pt}{2pt plus 4pt minus 2pt}{-2pt plus 2pt minus 2pt}

\makeatletter
\def\th@remark{%
  \thm@headfont{\bfseries}%
  \normalfont 
  \thm@preskip\topsep \divide\thm@preskip\tw@
  \thm@postskip\thm@preskip
}
\makeatother

\theoremstyle{definition}

\tcolorboxenvironment{theorem}{
  breakable,
  colback=black!10,
  colframe=white,
  width=\linewidth, 
  enlarge left by=0pt,
  enlarge right by=0pt,
  boxsep=5pt,
  boxrule=0pt,
  left=0pt,right=0pt,top=0pt,bottom=0pt,
  arc=8pt,
  before skip=\topsep,
  after skip=\topsep
}

\tcolorboxenvironment{lemma}{
  breakable,
  colback=black!10,
  colframe=white,
  width=\linewidth,
  enlarge left by=0pt,
  enlarge right by=0pt,
  boxsep=5pt,
  boxrule=0pt,
  left=0pt,right=0pt,top=0pt,bottom=0pt,
  arc=8pt,
  before skip=\topsep,
  after skip=\topsep
}

\tcolorboxenvironment{corollary}{
  breakable,
  colback=black!10,
  colframe=white,
  width=\linewidth,
  enlarge left by=0pt,
  enlarge right by=0pt,
  boxsep=5pt,
  boxrule=0pt,
  left=0pt,right=0pt,top=0pt,bottom=0pt,
  arc=8pt,
  before skip=\topsep,
  after skip=\topsep
}

\tcolorboxenvironment{proposition}{
  breakable,
  colback=black!10,
  colframe=white,
  width=\linewidth,
  enlarge left by=0pt,
  enlarge right by=0pt,
  boxsep=5pt,
  boxrule=0pt,
  left=0pt,right=0pt,top=0pt,bottom=0pt,
  arc=8pt,
  before skip=\topsep,
  after skip=\topsep
}

\newtheorem{definition}{Definition}[section]
\tcolorboxenvironment{definition}{
  breakable,
  colback=black!10,
  colframe=white,
  width=\linewidth,
  enlarge left by=0pt,
  enlarge right by=0pt,
  boxsep=5pt,
  boxrule=0pt,
  left=0pt,right=0pt,top=0pt,bottom=0pt,
  arc=8pt,
  before skip=\topsep,
  after skip=\topsep
}

\tcolorboxenvironment{assumption}{
  breakable,
  colback=black!10,
  colframe=white,
  width=\linewidth,
  enlarge left by=0pt,
  enlarge right by=0pt,
  boxsep=5pt,
  boxrule=0pt,
  left=0pt,right=0pt,top=0pt,bottom=0pt,
  arc=8pt,
  before skip=\topsep,
  after skip=\topsep
}

\tcolorboxenvironment{claim}{
  breakable,
  colback=black!10,
  colframe=white,
  width=\linewidth,
  enlarge left by=0pt,
  enlarge right by=0pt,
  boxsep=5pt,
  boxrule=0pt,
  left=0pt,right=0pt,top=0pt,bottom=0pt,
  arc=8pt,
  before skip=\topsep,
  after skip=\topsep
}

\tcolorboxenvironment{problem}{
  breakable,
  colback=black!10,
  colframe=white,
  width=\linewidth,
  enlarge left by=0pt,
  enlarge right by=0pt,
  boxsep=5pt,
  boxrule=0pt,
  left=0pt,right=0pt,top=0pt,bottom=0pt,
  arc=8pt,
  before skip=\topsep,
  after skip=\topsep
}

\tcolorboxenvironment{question}{
  breakable,
  colback=black!10,
  colframe=white,
  width=\linewidth,
  enlarge left by=0pt,
  enlarge right by=0pt,
  boxsep=5pt,
  boxrule=0pt,
  left=0pt,right=0pt,top=0pt,bottom=0pt,
  arc=8pt,
  before skip=\topsep,
  after skip=\topsep
}

\newtcolorbox{titleblock}{
  enhanced,
  frame hidden,
  colback=CarolinaUltraLight,
  colframe=CarolinaUltraLight,
  boxrule=0pt,
  arc=10pt,
  left=14pt,
  right=14pt,
  top=14pt,
  bottom=14pt,
  width=\linewidth,
  before skip=12pt plus 4pt,
  after skip=12pt plus 4pt,
  grow to left by=1.5pt,
  grow to right by=1.5pt,
  before upper={
    \setlength{\parindent}{0cm}
    \setlength{\parskip}{0.5cm}
  }
}

\crefname{theorem}{Theorem}{Theorems}
\crefname{proposition}{Proposition}{Propositions}
\crefname{lemma}{Lemma}{Lemmas}
\crefname{corollary}{Corollary}{Corollaries}
\crefname{definition}{Definition}{Definitions}
\crefname{assumption}{Assumption}{Assumptions}
\crefname{remark}{Remark}{Remarks}
\crefname{problem}{Problem}{Problems}
\crefname{property}{Property}{property}
\crefname{question}{Question}{Questions}

\numberwithin{equation}{section}
\numberwithin{theorem}{section}
\numberwithin{proposition}{section}
\numberwithin{definition}{section}
\numberwithin{lemma}{section}
\numberwithin{assumption}{section}
\numberwithin{remark}{section}

\newcommand\metadataformat[2][]{{\small {\bfseries #1:} #2}}

\def\1{\bm{1}}

\makeatletter
\let\save@mathaccent\mathaccent
\newcommand*\if@single[3]{%
    \setbox0\hbox{${\mathaccent"0362{#1}}^H$}%
    \setbox2\hbox{${\mathaccent"0362{\kern0pt#1}}^H$}%
    \ifdim\ht0=\ht2 #3\else #2\fi
}
\newcommand*\rel@kern[1]{\kern#1\dimexpr\macc@kerna}
\newcommand*\widebar[1]{\@ifnextchar^{{\wide@bar{#1}{0}}}{\wide@bar{#1}{1}}}
\newcommand*\wide@bar[2]{\if@single{#1}{\wide@bar@{#1}{#2}{1}}{\wide@bar@{#1}{#2}{2}}}
\newcommand*\wide@bar@[3]{%
    \begingroup
    \def\mathaccent##1##2{%
        \let\mathaccent\save@mathaccent
        \if#32 \let\macc@nucleus\first@char \fi
        \setbox\z@\hbox{$\macc@style{\macc@nucleus}_{}$}%
        \setbox\tw@\hbox{$\macc@style{\macc@nucleus}{}_{}$}%
        \dimen@\wd\tw@
        \advance\dimen@-\wd\z@
        \divide\dimen@ 3
        \@tempdima\wd\tw@
        \advance\@tempdima-\scriptspace
        \divide\@tempdima 10
        \advance\dimen@-\@tempdima
        \ifdim\dimen@>\z@ \dimen@0pt\fi
        \rel@kern{0.6}\kern-\dimen@
        \if#31
        \overline{\rel@kern{-0.6}\kern\dimen@\macc@nucleus\rel@kern{0.4}\kern\dimen@}%
        \advance\dimen@0.4\dimexpr\macc@kerna
        \let\final@kern#2%
        \ifdim\dimen@<\z@ \let\final@kern1\fi
        \if\final@kern1 \kern-\dimen@\fi
        \else
        \overline{\rel@kern{-0.6}\kern\dimen@#1}%
        \fi
    }%
    \macc@depth\@ne
    \let\math@bgroup\@empty \let\math@egroup\macc@set@skewchar
    \mathsurround\z@ \frozen@everymath{\mathgroup\macc@group\relax}%
    \macc@set@skewchar\relax
    \let\mathaccentV\macc@nested@a
    \if#31
    \macc@nested@a\relax111{#1}%
    \else
    \def\gobble@till@marker##1\endmarker{}%
    \futurelet\first@char\gobble@till@marker#1\endmarker
    \ifcat\noexpand\first@char A\else
    \def\first@char{}%
    \fi
    \macc@nested@a\relax111{\first@char}%
    \fi
    \endgroup
    }
\makeatother
\let\bar\widebar

\DeclareMathAlphabet{\mathsfit}{\encodingdefault}{\sfdefault}{m}{sl}
\SetMathAlphabet{\mathsfit}{bold}{\encodingdefault}{\sfdefault}{bx}{n}

\DeclareMathOperator*{\argmax}{arg\,max}

\let\tilde\widetilde
\let\hat\widehat

\usepackage{inconsolata}

\usepackage{ulem}
\usepackage{listings}
\crefname{definition}{Def.}{Defs.}

\begin{document}

\makeatletter
\def\blfootnote{\gdef\@thefnmark{}\@footnotetext}
\makeatother

\makeatletter
\pagestyle{fancy}
\fancyhf{}
\renewcommand{\headrulewidth}{1pt}
\chead{\small\bf Leave It to the Experts: Detecting Knowledge Distillation via MoE Expert Signatures
}
\cfoot{\thepage}
\thispagestyle{fancy}
\makeatother

\makeatletter
\def\icmldate#1{\gdef\@icmldate{#1}}
\icmldate{\today}
\makeatother

\makeatletter
\fancypagestyle{fancytitlepage}{
  \fancyhead{}
  \lhead{\includegraphics[height=0.8cm]{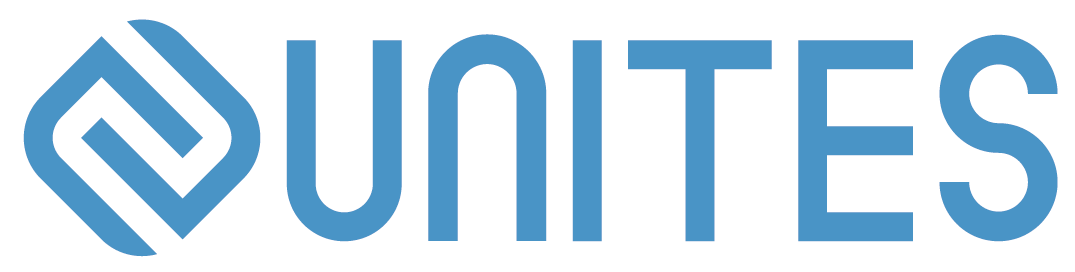}}
  \rhead{\it \@icmldate}
  \cfoot{}
}
\makeatother

\thispagestyle{fancytitlepage}

\vspace*{0.5em}

\noindent
\begin{titleblock}
    {\setlength{\parskip}{0cm}
     \raggedright
     {\setstretch{1.2}
      \LARGE\sffamily\bfseries
      Leave It to the Experts: Detecting Knowledge Distillation via MoE Expert Signatures
      \par}
    }
    \vskip 0.2cm
    
    \begin{icmlauthorlist}
\mbox{Pingzhi Li$^{\dagger\,1}$},
\mbox{Morris Yu-Chao Huang$^{\dagger\,1}$}, 
\mbox{Zhen Tan$^2$}, 
\mbox{Qingquan Song$^3$}, 
\mbox{Jie Peng$^1$},
\mbox{Kai Zou$^4$},
\mbox{Yu Cheng$^5$},
\mbox{Kaidi Xu$^6$},
and \mbox{Tianlong Chen$^1$}
\end{icmlauthorlist}

$^{1\,}$UNC-Chapel Hill 
\quad $^{2\,}$Arizona State University
\quad $^{3\,}$Individual Contributor
\quad $^{4\,}$NetMind.AI\\
\quad $^{5\,}$The Chinese University of Hong Kong
\quad $^{6\,}$City University of Hong Kong

$^{\dagger}$ Equal Contribution 
    \vskip 0.2cm
    
    Knowledge Distillation (KD) accelerates training of large language models (LLMs) but poses intellectual property protection and LLM diversity risks. Existing KD detection methods based on self-identity or output similarity can be easily evaded through prompt engineering. We present a KD detection framework effective in both white-box and black-box settings by exploiting an overlooked signal: the transfer of MoE ``structural habits'', especially internal routing patterns. Our approach analyzes how different experts specialize and collaborate across various inputs, creating distinctive fingerprints that persist through the distillation process. To extend beyond the white-box setup and MoE architectures, we further propose \texttt{Shadow-MoE}, a black-box method that constructs proxy MoE representations via auxiliary distillation to compare these patterns between arbitrary model pairs. We establish a comprehensive, reproducible benchmark that offers diverse distilled checkpoints and an extensible framework to facilitate future research. Extensive experiments demonstrate $>94\%$ detection accuracy across various scenarios and strong robustness to prompt-based evasion, outperforming existing baselines while highlighting the structural habits transfer in LLMs.
    
    \vskip 0.2cm
    {\setlength{\parskip}{0cm}
     \centering
     \makebox[\linewidth]{
        \metadataformat[Code]{
            \href{https://github.com/unites-lab/shadow-moe}{https://github.com/unites-lab/shadow-moe}
        }
     }
    }
\end{titleblock}

\blfootnote{%
$^{\textrm{\Letter}}$ Correspondence email: \{pingzhi, tianlong\}@cs.unc.edu
\\[2.5em]
\ifcsname @icmlpreprint\endcsname
  \textit{\csname @icmlpreprint\endcsname}%
\fi
}

\section{Introduction}
Knowledge Distillation~(KD)~\citep{hinton2015distillingknowledgeneuralnetwork} has emerged as a cornerstone technique for democratizing large language models~(LLMs), enabling the transfer of capabilities from computationally expensive and larger teacher models to more efficient and smaller student models. This paradigm has facilitated the training and deployment of powerful AI systems across resource-constrained environments~\citep{gou2021knowledge,wang2021knowledge,yang2025qwen3} and accelerated the development of specialized models for domain-specific applications~\citep{xu2024survey}. However, the widespread adoption of KD has introduced critical challenges to the LLM ecosystem: unauthorized distillation threatens intellectual property rights of model developers~\citep{maini2021dataset,li2025doge}, while excessive reliance on a few teacher models risks homogenizing the model landscape and stifling innovation~\citep{krishna2019thieves,qiu2025phybench}.

Detecting whether a model has undergone knowledge distillation is therefore crucial for both protecting commercial interests and understanding the provenance of AI systems. Existing detection approaches fall into two main categories: \textit{identity-based methods} that probe models' self-identity knowledge~\citep{lee2025quantificationlargelanguagemodel}, and \textit{behavior-based methods} that analyze output distribution similarities~\citep{mattern2023membership}. However, these methods exhibit critical limitations. Identity-based approaches can be trivially defeated through prompt engineering or fine-tuning that alters surface-level responses while preserving distilled knowledge. Behavior-based methods struggle with high false positive rates, as models trained on similar data naturally exhibit overlapping behaviors even without distillation~\citep{carlini2021extracting}.

Our work begins with a novel observation: knowledge distillation transfers not merely the functional mapping from inputs to outputs, but also the \textit{structural habits} of the teacher model, \textit{i.e.} the internal computational patterns and decision-making pathways that characterize how the model processes information. Particularly, in Mixture-of-Experts~(MoE) architectures~\citep{shazeer2017outrageouslylargeneuralnetworks,fedus2022switchtransformersscalingtrillion,jiang2024mixtralexperts}, these structural habits manifest as distinctive expert routing patterns: \textbf{expert specialization} of which experts activate for specific input types, and \textbf{expert collaboration} of how experts co-activate and cluster, that emerge during training. These routing signatures are deeply embedded in the model's architecture and persist through the distillation process, making them robust indicators of knowledge transfer. This leads to our key research question: \textit{Can we leverage the structural signatures inherited through knowledge distillation, particularly the expert routing patterns in MoE models, to reliably detect when distillation has occurred between models?}

Recognizing that not all models employ MoE architectures and some only provide API-based text output access, we further introduce \texttt{Shadow-MoE}, a black-box extension that enables KD detection between arbitrary model pairs. \texttt{Shadow-MoE} works by constructing proxy MoE representations of black-box models through further lightweight text-level distillation, \textit{i.e.} training a proxy MoE model to mimic the input-output behavior of target models, thereby exposing accessible routing patterns that preserve the structural habits inherited during knowledge transfer even when direct access to model internals is unavailable.

Our contributions and findings are summarized as follows: ($1$) We formalize the KD detection task and introduce MoE Expert Signatures (\textit{i.e.} expert specialization and collaboration), a novel detection method that leverages inherited structural habits in expert routing patterns to identify distillation relationships with accuracy up to $94\%$. ($2$) We propose \texttt{Shadow-MoE}, a black-box extension that enables KD detection between arbitrary black-box models by constructing analyzable proxy representations, broadening the applicability beyond MoE architectures and further improving the accuracy to $100\%$. ($3$) To our knowledge, we are the first to introduce a benchmark with reproducible experimental protocols and diverse checkpoints, providing the research community with essential infrastructure for advancing distillation detection research.
\section{Preliminary}

\paragraph{Setting.}
Let $\mathcal{X}$ denote the input space and $\mathcal{Y}$ the output space.
We consider two models: a \emph{suspected teacher} $f_T: \mathcal{X} \to \Delta(\mathcal{Y})$ and a \emph{suspected student} $f_S: \mathcal{X} \to \Delta(\mathcal{Y})$, where $\Delta(\mathcal{Y})$ denotes the probability simplex over $\mathcal{Y}$.
We assume black-box query access to both models.
Here we define the following Knowledge Distillation Set in \cref{def:kd-set}.
\begin{definition}[Knowledge Distillation Set]
\label{def:kd-set}
The knowledge distillation set $\mathrm{KD}(f_T)$ is defined as the set of all possible student model(s) $f_S$ distilled from the teacher model $f_T$:
\begin{align}
\mathrm{KD}(f_T) &\coloneqq \{f_S : \exists \mathcal{L}_{\mathrm{KD}}, \mathcal{D}_{\mathrm{train}} \nonumber\\
\text{ s.t. } f_S &= \arg\min_{f} \mathcal{L}_{\mathrm{KD}}(f, f_T; \mathcal{D}_{\mathrm{train}})\}
\end{align}
where $\mathcal{L}_{\mathrm{KD}}$ is any knowledge distillation loss (e.g., KL divergence, MSE on logits).
\end{definition}
With this, we can define the formulation of the studied knowledge distillation detection below.

\subsection{Problem Formulation}\label{sec:problem}
We consider a query distribution $\mathcal{Q}$ over $\mathcal{X}\times \mathcal{D}$, where $\mathcal{D}=\{1,\dots,D\}$ indexes semantic \emph{domains/tasks} (e.g., math, code, medical, etc.)\footnote{Domains and tasks are detailed in \cref{sec:experiments}.}.
Each sample $(x,d)\sim \mathcal{Q}$ consists of a prompt $x\in\mathcal{X}$ and domain label $d\in \mathcal{D}$.
We aim to test whether a suspected student $f_S$ has been distilled from a teacher $f_T$.
Formally, the \emph{knowledge-distillation detection} task is defined as a hypothesis test in \cref{def:kd}:
\begin{definition}[Knowledge Distillation Detection]
\label{def:kd}
We define the knowledge distillation detection task as a binary hypothesis test:
\begin{align*}
\textstyle H_1:~ f_S \in \mathrm{KD}(f_T)
\quad\text{vs.}\quad
H_0:~ f_S \notin \mathrm{KD}(f_T),
\end{align*}
where $\mathrm{KD}(f_T)$ denotes models obtained by distilling from $f_T$.
\end{definition}

\paragraph{\texttt{Shadow-MoE} Construction.}
Because many models are dense or API-limited, we cannot access their routing directly.
We therefore propose to construct the shadow proxies for $f_S$ and $f_T$ that mimic each model's input-output behavior and expose analyzable routing signals as detailed in \cref{def:shadow-moe}.
\begin{definition}[\texttt{Shadow-MoE} Proxy]
\label{def:shadow-moe}
A \texttt{Shadow-MoE} proxy $g: \mathcal{X} \to \Delta(\mathcal{Y})$ for model $f$ is a sparse MoE with $L$ layers and $E_\ell$ experts at layer $\ell$, trained via:
\begin{align*}
g^* = \arg\min_{g \in \mathcal{G}_{\text{MoE}}} \mathbb{E}_{x \sim \mathcal{Q}_\mathcal{X}} \left[\mathcal{L}_{\text{distill}}(g(x), f(x))\right] + \lambda \Omega(g)
\end{align*}
The load-balancing regularizer $\Omega(g)$ encourages balanced expert usage across a batch:
\begin{align}
\Omega(g) &= \sum_{\ell=1}^L E_\ell \sum_{i=1}^{E_\ell}
\!\left(\bar p^{(\ell)}_i - \tfrac{1}{E_\ell}\right)^{\!2},
\nonumber\\
\bar p^{(\ell)}_i &= \frac{1}{n}\sum_{m=1}^{n} p^{(\ell)}_i(x_m),
\end{align}
where $p^{(\ell)}_i \in \Delta^{E_\ell}$ is the softmax routing distribution at layer~$\ell$.
This term discourages expert collapse and promotes diverse routing behaviors, following existing works~\citep{fedus2022switchtransformersscalingtrillion,jiang2024mixtralexperts,deepseekai2025deepseekv3technicalreport}.
The load-balancing regularizer encourages each expert to receive a roughly equal fraction of tokens, preventing degenerate proxies where a few experts dominate.
\end{definition}

\begin{figure}[t]
    \centering
    \includegraphics[width=0.9\linewidth]{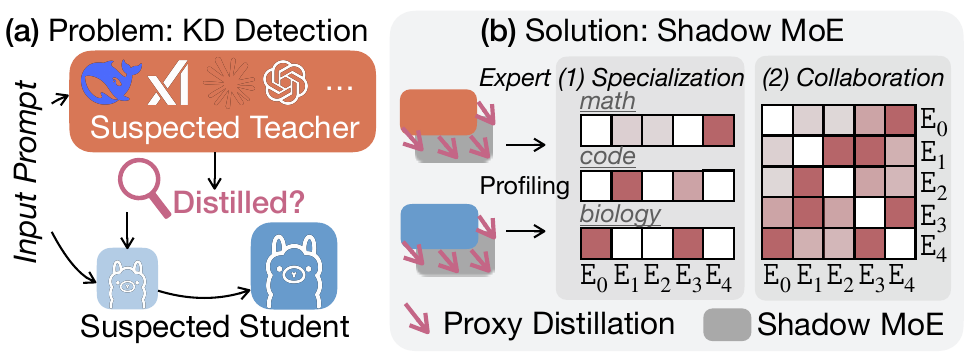}
    \vspace{-10pt}
    \caption{\textbf{Overview of our method.} \uline{\textbf{(a)}} Problem formulation: detecting whether a suspected student model was distilled from a teacher model, which is challenging when only black-box access is available. \uline{\textbf{(b)}} Our \texttt{Shadow-MoE} solution: we train proxy \texttt{Shadow-MoE} models to mimic both the suspected teacher or student, then analyze their expert routing patterns through two key measurements, \textit{i.e.} expert \textit{specialization}(task-specific activation profiles across different domains) and expert \textit{collaboration} (co-activation patterns between experts). Similar routing patterns between the shadow models provide evidence of a distillation relationship.}
    \label{fig:framework}
    \vspace{-12pt}
\end{figure}

\subsection{MoE Expert Specialization and Collaboration}\label{sec:special-collab}
Consider a sparse MoE model (or shadow proxy) $g$ with $L$ layers. At layer $\ell \in [L]$ with $E_\ell$ experts, let the the router outputs gating scores $p^{(\ell)}(x)\in\Delta^{E_\ell}$ and selects a top-$k_\ell$ set $\mathcal{K}^{(\ell)}(x)\subseteq\{1,\dots,E_\ell\}$.
Define the \emph{binary activation} for expert $i$:
\begin{align}\label{eq:effective_act_weight}
a^{(\ell)}_i(x)\;\coloneqq\;\mathbbm{1}\{\,i\in\mathcal{K}^{(\ell)}(x)\,\}\in\{0,1\}.
\end{align}

We identify two distinct signatures of MoE: Expert Specialization~\citep{li2023merge} and Expert Collaboration~\citep{luo2025occult,zhang2025advancingmoeefficiencycollaborationconstrained}.
Below are the definitions of two profiles.

\begin{definition}[Expert Specialization Profile]\label{def:special}
For domain $d\in[D]$ with $n_d$ queries and for layer $\ell$, define the empirical selection frequency
\begin{align}\label{eq:specialization}
\hat S^{(\ell)}_{\mathrm{bin},\,i,d}
~\coloneqq~
\frac{1}{n_d}\sum_{m:\,d_m=d} a^{(\ell)}_i(x_m).
\end{align}
To compare across domains with possibly varying $k_\ell(x)$, we normalize by the expected active expert count
\[
\hat\kappa^{(\ell)}_d
~=~
\frac{1}{n_d}\sum_{m:\,d_m=d} k_\ell(x_m),
\quad
\hat{\bar S}^{(\ell)}_{i,d}
~=~
\frac{\hat S^{(\ell)}_{\mathrm{bin},\,i,d}}{\hat\kappa^{(\ell)}_d},
\]
so that each column of $\hat{\bar S}^{(\ell)}$ sums to $1$. (If $k_\ell$ is constant, $\hat\kappa^{(\ell)}_d = k_\ell$.)
\end{definition}

\begin{definition}[Expert Collaboration Matrix]\label{def:collab}
At layer $\ell$, the empirical co-activation frequency between experts $i$ and $j$ is
\begin{align}\label{eq:collaboration}
\hat B^{(\ell)}_{i,j}
~\coloneqq~
\frac{1}{n}\sum_{m=1}^n a^{(\ell)}_i(x_m)\,a^{(\ell)}_j(x_m),
\quad i\neq j,
\end{align}
with $\hat B^{(\ell)}_{i,i}=0$. To obtain a probability-normalized version, let
\begin{align}
\widehat{\mathbb{E}}[k_\ell(k_\ell-1)]
~&=~
\frac{1}{n}\sum_{m=1}^n k_\ell(x_m)\big(k_\ell(x_m)-1\big),
\nonumber \\
\hat{\bar B}^{(\ell)}_{i,j}
~&=~
\frac{\hat B^{(\ell)}_{i,j}}{\widehat{\mathbb{E}}[k_\ell(k_\ell-1)]},
\end{align}
so that $\sum_{i\neq j}\hat{\bar B}^{(\ell)}_{i,j}=1$ and diagonal remains $0$.
\end{definition}
The specialization and collaboration profile from \cref{def:special,def:collab} are illustrated in \cref{fig:framework}.

\paragraph{Permutation Invariance.}
MoE expert labels are arbitrary; two models may differ by permutations yet implement the same routing function.
We thus compare specialization and collaboration signatures only via permutation-invariant distances.

\paragraph{Pair Classification Task.}
Given domain $d\in\{1,\dots,9\}$ (see \cref{sec:experiment-setup} for detail domain categories) and a pair of student checkpoints $\mathcal{S}_d=\{f_{S,d}^{\mathrm{KD}},\,f_{S,d}^{\mathrm{scratch}}\}$.
We define $f^{\mathrm{scratch}}$ as the model train from scratch without any supervision derived from $f_T$ (e.g., teacher-generated text, hidden states, or reward signals).
We cast KD detection as a paired binary classification problem in our experiments in \cref{sec:semi-black-box,sec:pure-black-box}:
The goal is to select the distilled model in each pair.
Specifically, each detector produces a scalar score $s(f_T,f_S)\in\mathbb{R}$, where larger values indicate a higher likelihood that $f_S$ is distilled from $f_T$.
For \texttt{Shadow-MoE}, we calculate the average of two signature: specialization $d_{\mathrm{spec}}$ and collaboration $d_{\mathrm{collab}}$ using the permutation-invariant Wasserstein distance in \eqref{eq:d_spec} and \eqref{eq:d_collab}.
Baselines (e.g. Idiosyncrasies~\citep{sun2025idiosyncrasieslargelanguagemodels}) provide their own monotone scores.
We report \emph{pairwaise accuracy} as $\mathrm{Acc} \;=\; \frac{1}{9}\sum_{d=1}^{9}\mathbbm{1}\!\left[\hat{\imath}_d=\mathrm{KD}\right]$ and \emph{decision margin} $m_d = s\!\left(f_T,f_{S,d}^{\mathrm{KD}}\right)-s\!\left(f_T,f_{S,d}^{\mathrm{scratch}}\right)$ as metric present in \cref{fig:semi-black-box-results,fig:pure-black-box-results}.

\section{Methodology}\label{sec:method}
\subsection{Proxy \texttt{Shadow-MoE} Training}
We consider the problem of detecting whether a suspected \emph{student} model $f_S$ has been distilled from a \emph{teacher} model $f_T$, under the black-box setting.
Our key idea is to compare their \emph{expert routing signatures}, which are invariant to expert index permutations and provide a stable characterization of model behavior.
Since many foundation models are not explicitly sparse MoEs, we construct \emph{shadow proxies} (\cref{def:shadow-moe}) by training sparse MoEs $g_T$ and $g_S$ to mimic $f_T$ and $f_S$ respectively on query-response data.
The detection problem then reduces to comparing the specialization and collaboration profiles of $g_T$ and $g_S$.

\subsection{MoE Signature Extraction}\label{sec:sig_extract}
For each \texttt{Shadow-MoE} $g$, we compute two profiles at the last layer $\ell$:
\begin{itemize}
    \item \textbf{Expert Specialization} (\cref{def:special}): domain-dependent activation frequencies normalized to probability distributions across experts.
    \item \textbf{Expert Collaboration} (\cref{def:collab}): normalized co-activation patterns between expert pairs.
\end{itemize}
These two metrics capture complementary aspects of expert behavior:
\emph{specialization} reflects how domains are partitioned across experts,
while \emph{collaboration} reflects how experts jointly contribute within the same domain.

Since expert indices are arbitrary, we measure signature similarity using permutation-invariant Wasserstein distances (\cref{sec:special-collab}).
Let $\Pi_{E_\ell}$ denote the set of all $E_\ell \times E_\ell$ permutation matrices.
For the $\ell$-th MoE layer, we define:
{\small
\begin{alignat}{2}
d_{\mathrm{spec}}^{(\ell)}
&=~ \min_{\Pi \in \Pi_{E_\ell}}
\frac{1}{D} \sum_{d=1}^{D}
W_1\!\big(\Pi \hat{\bar S}^{(\ell)}_T[:,d],~
\hat{\bar S}^{(\ell)}_S[:,d]\big), &
\label{eq:d_spec} \\[3pt]
d_{\mathrm{collab}}^{(\ell)}
&=~ \min_{\Pi \in \Pi_{E_\ell}}
\frac{1}{E_\ell} \sum_{i=1}^{E_\ell}
W_1\!\big((\Pi \hat{\bar B}^{(\ell)}_T \Pi^\top)[i,:],~
\hat{\bar B}^{(\ell)}_S[i,:]\big), &
\label{eq:d_collab}
\end{alignat}
}
where $W_1(\cdot,\cdot)$ denotes the Wasserstein-1 distance between normalized distributions.
In practice, we calculate these distances only at the last MoE layer to obtain overall specialization and collaboration distances, as deeper layer representations often demonstrate more prompt-specific information~\citep{chen2025sealsteerablereasoningcalibration,li2025quantmoebenchexaminingposttrainingquantization}.

\subsection{Distillation Detection}
We cast our distillation detection as a pair classification task.
For each domain $d$, we receive a candidate pair
$\mathcal{S}_d=\{f_{S,d}^{\mathrm{KD}}, f_{S,d}^{\mathrm{scratch}}\}$ and select the more likely distilled model by score comparison.
We aggregate the two distances by a simple average:
$
\mathrm{score}=-\frac{1}{2}\left(d_{\mathrm{spec}} + d_{\mathrm{collab}}\right),
$
so that higher scores indicate stronger evidence that $f_S$ was distilled from $f_T$.

In \cref{alg:detection}, we detail a paired KD detection procedure.
Given a teacher $f_T$ and a candidate pair $\{f_S^{(1)}, f_S^{(2)}\}$, we query all models on a shared prompt set sampled from $\mathcal{Q}$. If the teacher or a student is non-MoE or API-limited, we train lightweight \texttt{Shadow-MoE} proxies $(g_T, g_S^{(1)}, g_S^{(2)})$ via \Cref{def:shadow-moe} to expose analyzable routing signals.
We then extract expert specialization and collaboration signatures $\Phi(g_T)$ and $\Phi(g_S^{(i)})$, compute the permutation-invariant Wasserstein distances $d_{\mathrm{spec}}$ and $d_{\mathrm{collab}}$ (Eqs.~\eqref{eq:d_spec}, \eqref{eq:d_collab}), and form a single score $s_i = -\tfrac{1}{2}\,(d_{\mathrm{spec}} + d_{\mathrm{collab}})$.
The predicted distilled model is $\hat{\imath} = \arg\max_{i\in\{1,2\}} s_i$. Larger scores indicate closer routing similarity to the teacher; we evaluate using pairwise accuracy and decision margins across domains in \cref{sec:experiments}.

\begin{algorithm}[H]
\caption{MoE Expert Signature Detection}
\label{alg:detection}
\begin{algorithmic}[1]
\Require Teacher $f_T$; student pair $\mathcal{S}=\{f_S^{(1)}, f_S^{(2)}\}$; query budget $n$
\Ensure Predicted index $\hat{\imath}\in\{1,2\}$
\State Sample $\{(x_m, d_m)\}_{m=1}^n \sim \mathcal{Q}$ \Comment{shared prompts}
\If {teacher or any student is non-MoE or API-limited}
    \State Train proxy $g_T$ to mimic $f_T$ via Def.~\ref{def:shadow-moe}
    \For{each $f_S^{(i)}\in\mathcal{S}$}
        \State If non-MoE/API-limited, train proxy $g_S^{(i)}$; else set $g_S^{(i)}\gets f_S^{(i)}$
    \EndFor
\Else
    \State $g_T \gets f_T$; $g_S^{(i)} \gets f_S^{(i)}$ for $i\in\{1,2\}$
\EndIf
\For{$i \in \{1,2\}$}
    \State Extract signatures $\Phi(g_T)$ and $\Phi(g_S^{(i)})$
    \State Compute $d_{\mathrm{spec}}, d_{\mathrm{collab}}$ via \eqref{eq:d_spec}, \eqref{eq:d_collab}
    \State Score: $s_i \gets -\tfrac{1}{2}\big(d_{\mathrm{spec}} + d_{\mathrm{collab}}\big)$
\EndFor
\State \Return $\hat{\imath} \gets \arg\max_{i\in\{1,2\}} s_i$
\end{algorithmic}
\end{algorithm}

\begin{figure}[htbp]
    \centering
    \includegraphics[width=1\linewidth]{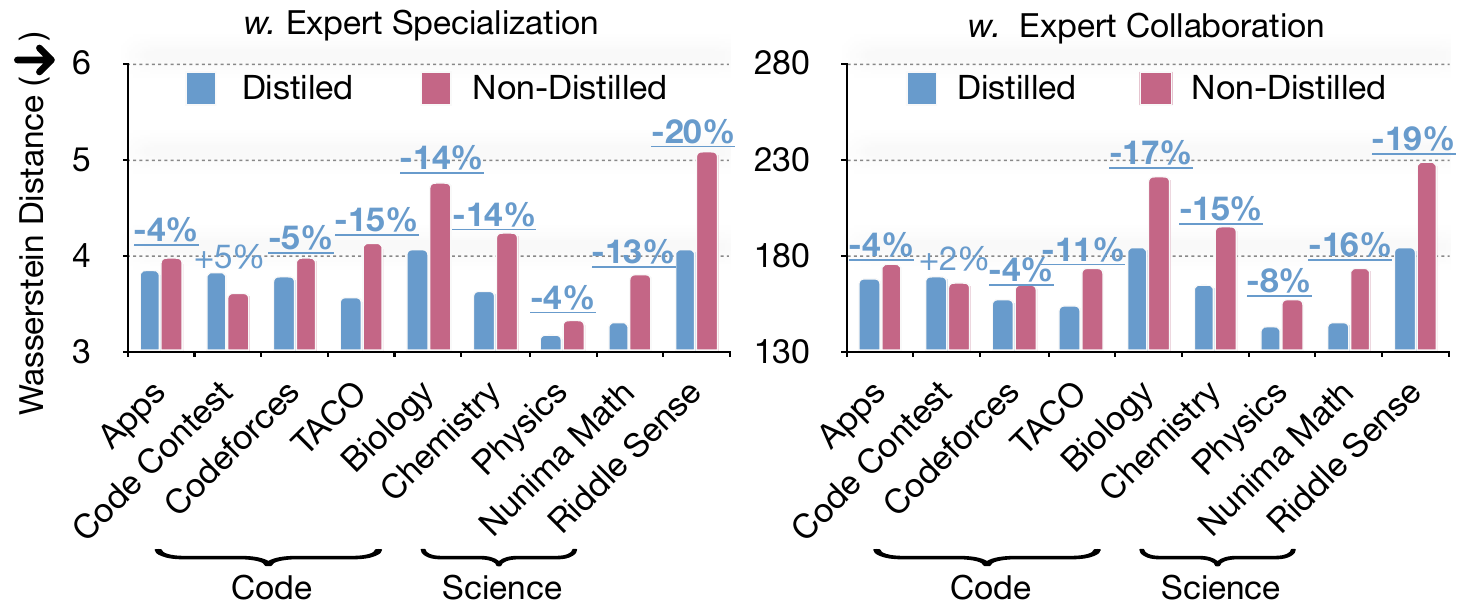}
    \vspace{-20pt}
    \caption{\textbf{Predicted scores with the black-box teachers and white-box students setting of \texttt{Shadow-MoE}.} We show Wasserstein distances between the teacher's Shadow-MoE proxy and student models for both Expert Specialization (left) and Expert Collaboration (right) metrics. Blue bars represent distilled students, while pink bars represent non-distilled students trained from scratch. Percentage differences indicate the relative reduction in distance for distilled models compared to their non-distilled counterparts. Successfully detected tasks (where distilled models show lower distances than non-distilled) are marked with \underline{\textbf{bold underline}}. Lower distances indicate stronger routing signature similarity, providing evidence of knowledge distillation.}
    \label{fig:semi-black-box-results}
    \includegraphics[width=1\linewidth]{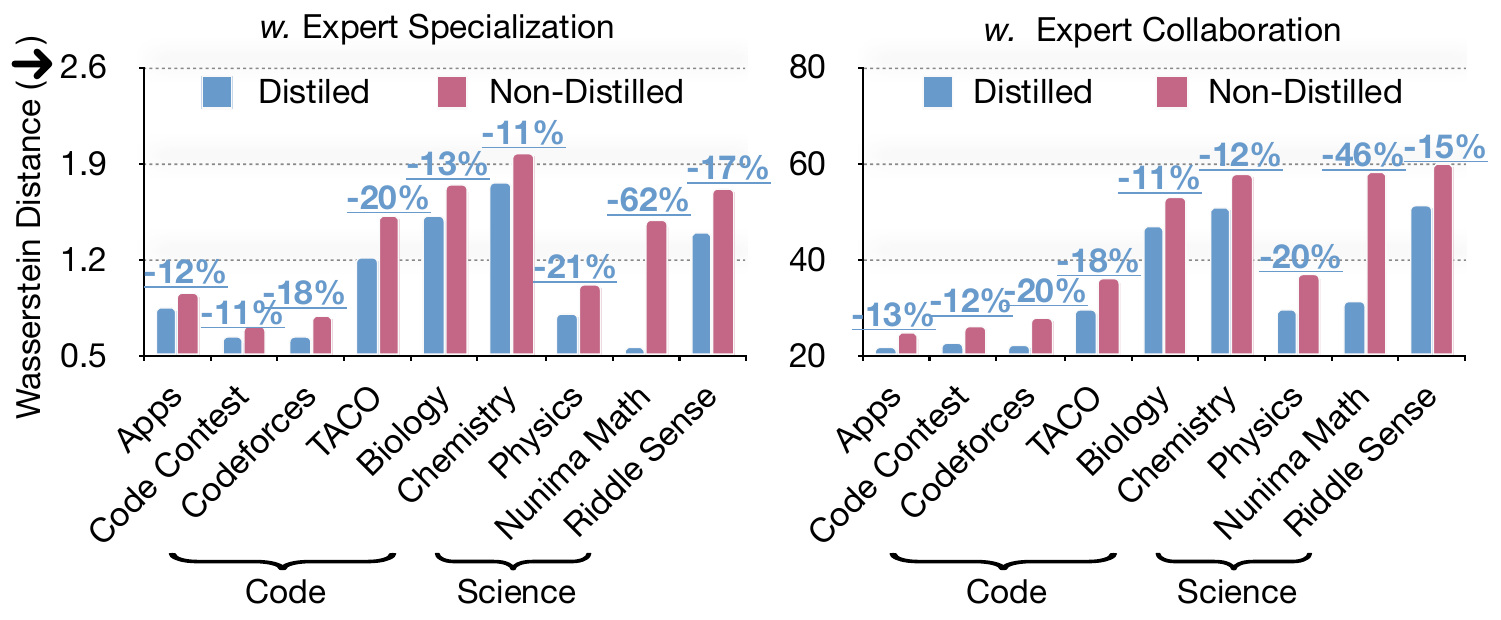}
    \vspace{-20pt}
    \caption{\textbf{Predicted scores with the black-box teachers and black-box students setting of \texttt{Shadow-MoE}.} Same metrics as Figure~\ref{fig:semi-black-box-results}, but with \texttt{Shadow-MoE} proxies constructed for both teacher and student models. Despite the additional proxy approximation for students, the method maintains even stronger detection performance with $100\%$ accuracy between distilled (blue) and non-distilled (pink) models across all tasks.}
    \label{fig:pure-black-box-results}
    \vspace{-15pt}
\end{figure}

\section{Experiments}\label{sec:experiments}

\subsection{Experimental Setup}\label{sec:experiment-setup}

\paragraph{Calibration Dataset.} We construct our calibration dataset by randomly sampling $280$ prompts from the \texttt{allenai/tulu-3-sft-mixture} dataset~\citep{lambert2024tulu3}, which provides diverse task coverage across multiple domains, including mathematics, coding, and general reasoning. This prompt set serves two purposes in our pipeline: \ding{182}~Training \texttt{Shadow-MoE} proxies via distillation to mimic the input-output behavior of suspected teacher and student models (\cref{def:shadow-moe}); \ding{183}~Profiling expert routing patterns to extract specialization and collaboration signatures for detection (\cref{def:special,def:collab}). The moderate dataset size provides a sweet spot between computational efficiency and sufficient coverage to capture representative routing behaviors across domains.

\begin{wraptable}{r}{0.6\textwidth}
      \centering
      \vspace{-10pt}
      \caption{Configuration of the LLMs used in this work.}
      \vspace{-8pt}
       \resizebox{1\linewidth}{!}{
        \begin{tabular}{r|ccc|c}
        \toprule
        \midrule
        \textbf{Model} & \textbf{Top-K} & \textbf{\# Shared Experts} & \textbf{\# Routed Experts} & \textbf{Model Size} \\
        \midrule
        \texttt{DeepSeek-R1} & $8$ & $1$ & $256$ & $685$B \\
        \texttt{Moonlight-16B-A3B} & $6$ & $2$ & $64$ & $16$B \\
        \texttt{OLMoE-1B-7B} & $8$ & $0$ & $64$ & $7$B \\
        \midrule
        \bottomrule
        \end{tabular}}
      \label{tab:model-config}
      \vspace{-12pt}
\end{wraptable}

\paragraph{Model Preparation.} We employ DeepSeek-R1~\citep{guo2025deepseek} as our black-box teacher model, to which we only have access to text outputs without internal information. To construct analyzable proxy representations, we train Moonlight-16B-A3B~\citep{liu2025muonscalablellmtraining} as the shadow MoE model using the calibration dataset to mimic the teacher's input-output behavior. For student model evaluation, we use OLMoE-1B-7B~\citep{muennighoff2024olmoeopenmixtureofexpertslanguage} as the candidate architecture and train it under two conditions, with and without distillation, across $9$ domain-specific datasets spanning four categories: \textit{Code} (\texttt{TACO}, \texttt{Apps}, \texttt{Code Contests}, \texttt{Codeforces}), \textit{Math} (\texttt{NuminaMath}), \textit{Science} (\texttt{Chemistry}, \texttt{Biology}, \texttt{Physics}), and \textit{Puzzle} (\texttt{Riddle\ Sense}). This yields $18$ student checkpoints ($9$ datasets $\times$ $2$ training conditions), enabling comprehensive evaluation of our detection method across diverse domain specializations. Given the two student checkpoints of each dataset, we will apply the baseline methods and our \texttt{Shadow-MoE} to predict which one is distilled from the suspected teacher, as a binary classification task. The configuration of LLMs used in our experiments is presented in Table~\ref{tab:model-config}.

\paragraph{Detection Baselines.} To validate the effectiveness of our method, we adopt the following baselines for comparison: (1)~\textit{Linear model embedding} that extracts response embeddings from candidate models and calculate the cosine similarity between them as distillation score; (2)~\textit{BERT embedding} that uses \texttt{ModernBERT-base}~\citep{modernbert}, a modern BERT-style model, to encode the response from candidate models and calculate the cosine similarity between them as distillation score; (3)~\textit{Idiosyncrasies}~\citep{sun2025idiosyncrasieslargelanguagemodels} that leverages fine-tuned text embedding models~(\textit{i.e.} LLM2vec) to identify the output patterns across different candidate LLMs by training on \textit{held-out} teacher-generated responses, \textit{i.e.} the calibration dataset in our setting; (4)~\textit{Model self-identity}~\citep{lee2025quantificationlargelanguagemodel} that employs jailbreaking techniques, \textit{i.e.}, GPTFuzz~\citep{yu2024gptfuzzerredteaminglarge}, to probe for identity consistency contradictions, detecting whether a suspected student model inadvertently reveals knowledge of the teacher model's identity through adversarial prompting. The first two baselines rely on surface-level text representations, while the latter two capture behavioral and identity-related signals that may indicate distillation relationships.

\vspace{-4pt}
\subsection{White-box Students, Black-box Teachers}\label{sec:semi-black-box}

\begin{wraptable}{r}{0.6\textwidth}
      \centering
      \vspace{-5pt}
      \caption{Classification accuracies of various methods in \textit{white-box students, black-box teacher}s setting. We mark the highest accuracy for each task set with \textbf{bold}.}
      \vspace{-8pt}
      \newcolumntype{H}{>{\columncolor{gray!10}}c}
       \resizebox{1\linewidth}{!}{
        \begin{tabular}{r|ccccH}
        \toprule
        \midrule
        \textbf{Task Set} & \textbf{Linear} & \textbf{BERT} & \textbf{Idiosyncrasies} & \textbf{Self-Identify} & \textbf{\texttt{Shadow-MoE}} \\
        \midrule
        \textit{Code} & $50\%$ & $50\%$ & $50\%$ & $0\%$ & $\mathbf{75\%}$  \\
        \textit{Math} & $\mathbf{100\%}$ & $\mathbf{100\%}$ & $\mathbf{100\%}$ & $0\%$ & $\mathbf{100\%}$  \\
        \textit{Science} & $33\%$ & $67\%$ & $\mathbf{100\%}$ & $0\%$ & $\mathbf{100\%}$  \\
        \textit{Puzzle} & $0\%$ & $\mathbf{100\%}$ & $\mathbf{100\%}$ & $0\%$ & $\mathbf{100\%}$  \\
        \midrule
        \midrule
        Average & $46\%$ & $54\%$ & $88\%$ & $0\%$ & $\mathbf{94\%}$  \\
        \bottomrule
        \end{tabular}}
      \label{tab:semi-black-box-baseline}
      \vspace{-15pt}
\end{wraptable}
\paragraph{Setting.} We first evaluate our \texttt{Shadow-MoE} on a semi-black-box setting, where we have black-box access to the suspected teacher LLMs while white-box access to the suspected student MoE LLMs. Specifically, we construct Shadow-MoE proxies only for the black-box teacher (DeepSeek-R1) using the calibration dataset of $280$ prompts, training Moonlight-16B-A3B via text-level distillation for $3$ epochs with a learning rate of $5 \times 10^{-6}$. For student models, we directly extract routing patterns from the white-box OLMoE-1B-7B checkpoints without requiring proxy construction. Each task set consists of both distilled and non-distilled student models trained on domain-specific data, creating a binary classification problem where we test whether the distilled students align more closely with the teacher than their non-distilled counterparts, and compare baseline methods with ours.

\vspace{-5pt}
\paragraph{Superior distillation detection performance of \texttt{Shadow-MoE}.} Our method achieves an average accuracy of $94\%$ across all task sets, substantially outperforming conventional embedding-based approaches. The performance is particularly strong on Math, Science, and Puzzle tasks, where we achieve $100\%$ accuracy. Notably, the self-identity baseline completely fails, with $0\%$ across all tasks, demonstrating that prompt-based identity probing cannot reliably detect structural knowledge transfer when models are fine-tuned on domain-specific data without identity knowledge.

\vspace{-5pt}
\paragraph{Consistent separation between distilled and non-distilled models via routing signatures.} Figure~\ref{fig:semi-black-box-results} demonstrates the discriminative effectiveness of our \texttt{Shadow-MoE} approach across diverse domains. Distilled models consistently exhibit lower Wasserstein distances to the teacher's proxy compared to their non-distilled counterparts, with reductions ranging from $4\%$ to $20\%$ for Expert Specialization and $2\%$ to $19\%$ for Expert Collaboration. This pattern holds across all evaluated tasks except for \textit{Code Contest}, where the non-distilled model shows $5\%$ and $2\%$ lower distance, likely due to the code domain inducing similar response structures even without explicit distillation. The complementary nature of the two metrics, with Expert Specialization capturing domain-specific routing preferences and Expert Collaboration revealing inter-expert dependencies, provides echoing evidence for detecting knowledge transfer relationships.

\paragraph{Idiosyncrasies as a competitive baseline.} The Idiosyncrasies approach emerges as the strongest one among existing baselines with $88\%$ average accuracy. This method, which trains a text embedding model (\textit{i.e.}, ModernBERT-base) to identify output patterns specific to different LLMs, captures surface-level stylistic signatures that persist through distillation. However, it shows limitations on Code tasks ($50\%$ accuracy) where domain-specific syntax and conventions may dominate over model-specific patterns, while routing patterns used in \texttt{Shadow-MoE} provide more consistent signals across diverse domains.

\subsection{Black-box Students, Black-box Teachers}\label{sec:pure-black-box}
\vspace{-3pt}

\begin{wraptable}{r}{0.6\textwidth}
      \centering
      \vspace{-2pt}
      \caption{Classification accuracies of various methods in \textit{black-box students, black-box teacher}s setting. We mark the highest accuracy for each task set with \textbf{bold}. The \textit{Linear} baseline, requiring access to hidden states of suspected student models, is not available at this setting.}
      \vspace{-8pt}
      \newcolumntype{H}{>{\columncolor{gray!10}}c}
       \resizebox{1\linewidth}{!}{
        \begin{tabular}{r|ccccH}
        \toprule
        \midrule
        \textbf{Task Set} & \textbf{Linear} & \textbf{BERT} & \textbf{Idiosyncrasies} & \textbf{Self-Identify} & \textbf{\texttt{Shadow-MoE}} \\
        \midrule
        \textit{Code} & - & $50\%$ & $50\%$ & $0\%$ & $\mathbf{100\%}$  \\
        \textit{Math} & - & $\mathbf{100\%}$ & $\mathbf{100\%}$ & $0\%$ & $\mathbf{100\%}$  \\
        \textit{Science} & - & $67\%$ & $\mathbf{100\%}$ & $0\%$ & $\mathbf{100\%}$  \\
        \textit{Puzzle} & - & $\mathbf{100\%}$ & $\mathbf{100\%}$ & $0\%$ & $\mathbf{100\%}$  \\
        \midrule
        \midrule
        Average & - & $54\%$ & $88\%$ & $0\%$ & $\mathbf{100\%}$  \\
        \bottomrule
        \end{tabular}}
      \label{tab:pure-black-box-baseline}
      \vspace{-10pt}
\end{wraptable}

\paragraph{Setting.} We extend our evaluation to the most challenging pure black-box setting, where we have only output text access to both the suspected teacher and student models. Unlike Section~\ref{sec:semi-black-box} where we could directly extract routing patterns from white-box student MoE models, here we must construct \texttt{Shadow-MoE} proxies for both sides of the detection problem. Specifically, we train \texttt{Shadow-MoE} proxies for both the black-box teacher (DeepSeek-R1) and the black-box student models (OLMoE-1B-7B checkpoints) using the same calibration dataset and training configuration, \textit{i.e.} Moonlight-16B-A3B trained via text-level distillation for $3$ epochs with a learning rate of $5 \times 10^{-6}$. This introduces an additional layer of approximation for the student models, as we now compare proxy-to-proxy routing signatures rather than proxy-to-actual signatures.

\paragraph{Further improved distillation detection performance of \texttt{Shadow-MoE} in pure black-box setting.}
Remarkably, our method achieves perfect detection accuracy of $100\%$ across all task sets in the pure black-box setting, as shown in Table~\ref{tab:pure-black-box-baseline}, even surpassing its already strong performance in the semi-black-box setting. Figure~\ref{fig:pure-black-box-results} reveals more pronounced separation between distilled and non-distilled models compared to the white-box student setting, with Wasserstein distance reductions ranging from $11\%$ to $62\%$ for Expert Specialization and $11\%$ to $46\%$ for Expert Collaboration. Notably, even the previously challenging Code Contest task now shows clear separation with $11\%$ and $12\%$ lower distances for the distilled model. This superior performance suggests that \texttt{Shadow-MoE} achieves more precise distillation detection when investing additional computational resources to train proxy models for both teacher and student, likely benefiting from using the same pre-trained model architecture (Moonlight-16B-A3B) as the proxy for both sides.

\subsection{Ablation Study and Extended Analysis}\label{sec:ablation-study}

\begin{wrapfigure}{r}{0.5\textwidth}
    \centering
    \vspace{-20pt}
    \includegraphics[width=1\linewidth]{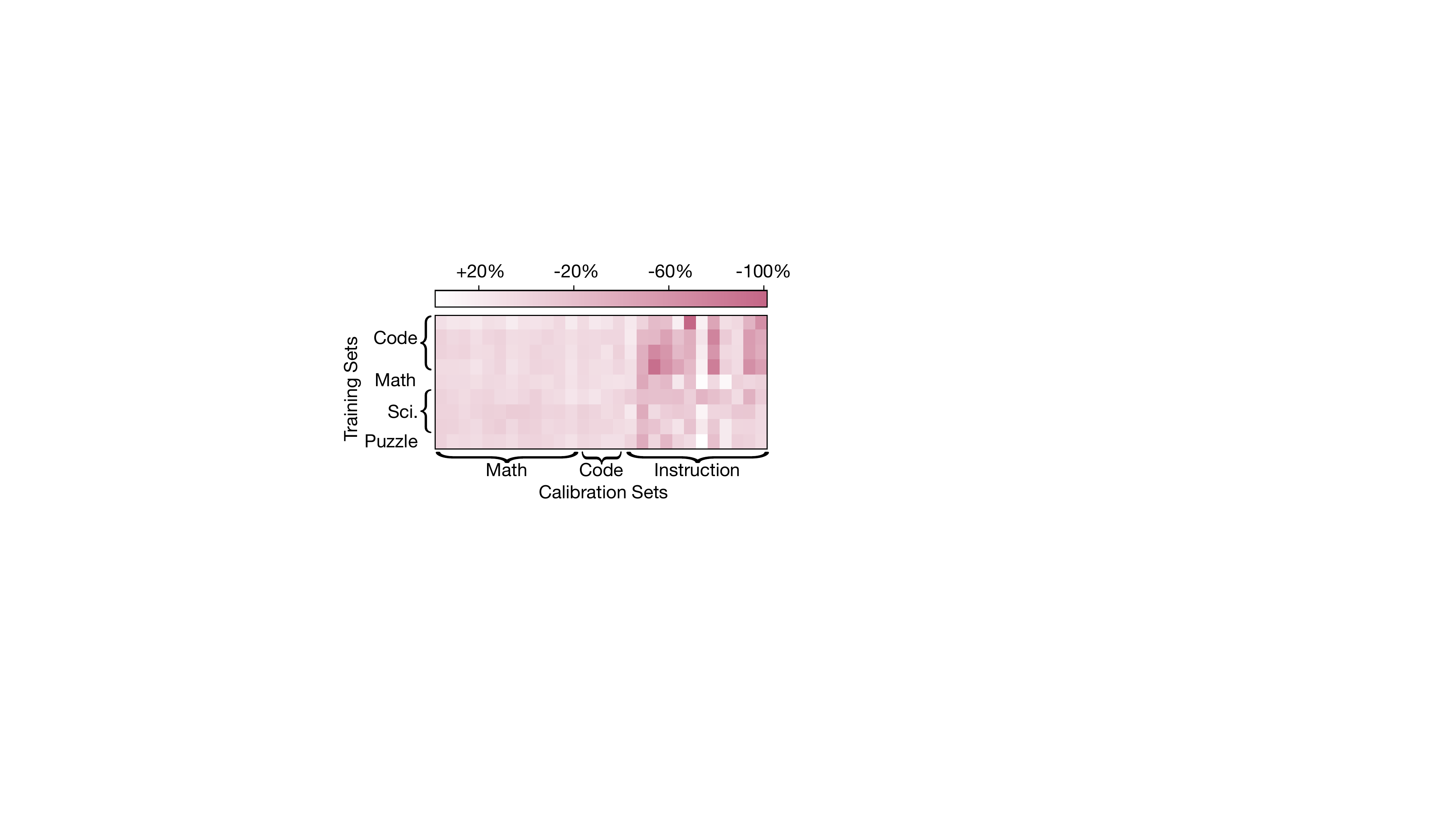}
    \vspace{-22pt}
    \caption{Relative Wasserstein distance reduction for distilled models compared to non-distilled models across different training and calibration set combinations. Darker colors indicate larger reductions (stronger detection signals), with percentages showing how much lower the distilled model's distance is relative to the non-distilled model. }
    \label{fig:task-transfer}
    \vspace{-18pt}
\end{wrapfigure}
\paragraph{Routing Pattern Transferability across Different Distillation and Calibration Tasks.} We investigate whether routing signatures remain discriminative when extracted using different calibration prompt sets than those used during training. We evaluate all $9$ training tasks against $28$ diverse calibration subsets sampled from various domains within the \texttt{allenai/tulu-3-sft-mixture} dataset. As shown in Figure~\ref{fig:task-transfer}, we measure the relative reduction in Wasserstein distance between distilled and non-distilled models, where more negative values (darker colors) indicate stronger detection signals. Surprisingly, specialized math and code calibration datasets fail to capture significant routing differences even for their corresponding training domains, showing only modest reductions. In contrast, general instruction-following calibration sets consistently achieve strong discriminative power across all task categories, with reductions reaching $-60\%$ to $-100\%$. This counterintuitive finding likely suggests that the most informative routing pattern changes induced by distillation occur in the processing of instruction-related tokens rather than domain-specific content.

\begin{wraptable}{r}{0.6\textwidth}
      \centering
      \caption{Ablation study on layer selection for routing signature extraction in the white-box students, black-box teachers setting.}
      \vspace{-10pt}
      \newcolumntype{H}{>{\columncolor{gray!10}}c}
       \resizebox{\linewidth}{!}{
        \begin{tabular}{r|cccH}
        \toprule
        \midrule
        \textbf{Task Set} & \textbf{First Layer} & \textbf{Median Layer} & \textbf{Last Layer (Ours)} \\
        \midrule
        \textit{Code} & $50\%$ & $75\%$ & $\mathbf{75\%}$  \\
        \textit{Math} & $\mathbf{100\%}$ & $\mathbf{100\%}$ & $\mathbf{100\%}$  \\
        \textit{Science} & $33\%$ & $67\%$ & $\mathbf{100\%}$  \\
        \textit{Puzzle} & $0\%$ & $\mathbf{100\%}$ & $\mathbf{100\%}$  \\
        \midrule
        \midrule
        Average & $46\%$ & $85\%$ & $\mathbf{94\%}$  \\
        \bottomrule
        \end{tabular}}
      \label{tab:layer-ablation}
      \vspace{-15pt}
\end{wraptable}
\paragraph{Routing Efficacy of Different MoE Layers.}
To validate our choice of using the last MoE layer for signature extraction, we conduct an ablation study comparing routing patterns from different layers in the semi-black-box setting (Section~\ref{sec:semi-black-box}). We extract expert specialization and collaboration signatures from three positions: \{the first, the median, and the last\} MoE layer. Table~\ref{tab:layer-ablation} presents the detection accuracy across different task sets. The results demonstrate that deeper layers provide increasingly discriminative routing signatures, with the last layer achieving the highest accuracy of $94\%$. The first layer shows nearly random discriminative power with $48\%$ accuracy, likely because early routing decisions are more influenced by surface-level token features rather than semantic content. This validates our design choice of using the final layer's routing patterns.

\section{Related Works}

\noindent\textbf{Mixture-of-Experts~(MoE)}~\citep{shazeer2017outrageouslylargeneuralnetworks} has shown promising results for efficiently scaling model capacity without a proportional increase in computational cost. This is typically achieved by replacing dense feed-forward layers with sparse MoE layers, where a routing mechanism directs each input token to a small subset of experts. Switch Transformers~\citep{fedus2022switchtransformersscalingtrillion} simplified MoE routing (\textit{i.e.}, top-$1$ routing) and demonstrated significant pre-training speedups and scalability up to trillion parameters by reducing communication and computational overheads. Mixtral-8x7B~\citep{jiang2024mixtralexperts} activates only two experts per token per layer but accesses a much larger total parameter count, illustrating that MoE can match the performance of equivalent full-parameter LLMs while utilizing far fewer active parameters. DeepSeek-MoE~\citep{dai2024deepseekmoeultimateexpertspecialization,deepseekai2025deepseekv3technicalreport} refined this architecture with fine-grained expert segmentation and shared experts, aiming for enhanced expert specialization and parameter efficiency. Moreover, \textit{expert specialization} naturally emerges as the gating network learns to route specific types of inputs to particular experts, reinforcing their proficiency~\citep{dai2024deepseekmoeultimateexpertspecialization,li2024mergecompressdemystifyefficient,wei2024skyworkmoedeepdivetraining}. \textit{Expert collaboration} refers to the co-activation of multiple experts to process certain input tokens, recently enabling reduced communication overhead and efficient expert parallelism through optimized expert placement and routing strategies~\citep{anonymous2025occult,zhang2025advancingmoeefficiencycollaborationconstrained}. In this work, we leverage expert specialization and collaboration as the underlying functional similarity inherited through distillation for detecting knowledge distillation.

\noindent\textbf{Knowledge Distillation~(KD)}~\citep{hinton2015distillingknowledgeneuralnetwork} has been a widely adopted model compression technique where a smaller ``student'' model is trained to replicate the behavior and inherit the capabilities of a larger, more powerful ``teacher'' model, to produce efficient yet powerful models~\citep{hsieh2023distilling,ma2021undistillablemakingnastyteacher,ma2022stingy,sanh2019distilbert}. In the context of LLMs, KD is usually performed at three levels of granularity, including: ($1$) \textit{layer hidden states-level} KD for aligning the student's intermediate hidden state representations with those of the teacher~\citep{chang2022distilhubert,liang2023less,lin2023lad}, ($2$) \textit{logits-level} KD for matching the teacher's final output probability distributions over tokens~\citep{anshumann2025sparselogitsamplingaccelerating,li2024mergecompressdemystifyefficient,yang2024survey}, and ($3$) \textit{output text-level} KD for replicating the teacher's generated text~\citep{bercovich2025llama,muennighoff2025s1,savani2025antidistillationsampling}. In this work, we focus on the most widely adopted \textbf{output text-level} KD as it is flexible to different student-teacher vocabularies or even black-box models with only API access, and produces minimal computing overhead~\citep{guo2025deepseek,muennighoff2025s1}. Recently, KD has gathered significant attention due to the rich semantic information in LLM reasoning traces, which has proven highly effective for transferring complex problem-solving abilities~\citep{guo2025deepseek,bercovich2025llama,muennighoff2025s1,savani2025antidistillationsampling}. However, it raises critical concerns about intellectual property protection and model homogenization\citep{savani2025antidistillationsampling}. Therefore, there is a growing need to quantify the extent of distillation and develop effective methods to detect if a model has been distilled from another~\citep{lee2025quantificationlargelanguagemodel}.

\noindent\textbf{Tracing LLMs to training data} coalesce around memorization/extraction, contamination/deduplication, and training-data attribution. Black-box extraction attacks show that individual training sequences can be recovered from deployed LMs and that vulnerability scales with model size~\citep{carlini2021extracting}. Follow-up measurement work quantifies how memorization grows with model capacity, duplication, and prompt context length~\citep{carlini2023quantifying}. To curb regurgitation and evaluation inflation, deduplication reduces verbatim emission and train–test overlap~\citep{lee2022deduplicating} and directly mitigates extraction risk~\citep{kandpal2022deduplicating}. Beyond aggregate leakage, \citet{akyurek2022towards} formalize fact tracing, retrieving ``proponent" training examples for generated assertions, and find that popular gradient- and embedding-based methods still lag strong IR baselines. For scalable per-example attribution, gradient-tracing via TracIn~\citep{pruthi2020tracin} and randomly projected after-kernel scoring via TRAK~\citep{park2023trak} estimate pointwise influence and scale to modern LLMs and CLIP-style VLMs. Collectively, these works motivate provenance-aware analyses when linking behaviors to pretraining corpora; in contrast, our paper pivots to \textit{model-internal} signals, which use MoE routing patterns as fingerprints to detect knowledge distillation relationships.
\section{Conclusion}\label{sec:conclusion}
We introduce a practical framework for detecting knowledge distillation that leverages Mixture-of-Experts routing signatures as structural fingerprints of model behavior.
Our approach rests on two key ideas:
(i) distillation transfers not only surface behavior but also structural habits in computation, and
(ii) these habits can be exposed and compared through lightweight \texttt{Shadow-MoE} proxies even in black-box settings.
Concretely, we defined two complementary routing profiles, \textit{i.e.} \emph{expert specialization} and \emph{expert collaboration}, and compared them via permutation-invariant \emph{Wasserstein} distances for distillation detection.
Across semi--black-box (\textit{i.e.} black-box teachers and white-box MoE students) and pure black-box (\textit{i.e.} black-box teachers and black-box students) settings, our method consistently outperforms embedding- and identity-based baselines, achieving high accuracy across diverse domains.
We release the benchmark with distilled and non-distilled checkpoints to facilitate future study.
We see this work as a step toward structure-aware alignment and defenses (\textit{e.g.}, structural watermarks, routing randomization).

\section*{Limitations}
Our results suggest that structural fingerprints provide a promising path toward provenance analysis for modern LLMs, complementing existing approaches based on identity prompts, text embeddings, or membership signals.
Looking ahead, we see three natural directions:
\textbf{Beyond MoE and richer structure} by extending signature to dense model and incorporate additional structure cues (\textit{e.g.} attention head usage).
\textbf{Alternative distillation channels} for detecting reward-model-mediated or RL-based distillation.
\textbf{Stronger guarantees and defenses} by exploring defensive mechanisms (\textit{e.g.} structural watermarks or routing randomization) to deter unauthorized distillation.
\section*{Acknowledgments}

Pingzhi Li, Morris Yu-Chao Huang, and Tianlong Chen are partially supported by Amazon Research Award and Cisco Faculty Award.

\bibliography{999_reference}

\begin{thebibliography}{57}
\providecommand{\natexlab}[1]{#1}
\providecommand{\url}[1]{\texttt{#1}}
\expandafter\ifx\csname urlstyle\endcsname\relax
  \providecommand{\doi}[1]{doi: #1}\else
  \providecommand{\doi}{doi: \begingroup \urlstyle{rm}\Url}\fi

\bibitem[Aky{\"u}rek et~al.(2022)Aky{\"u}rek, Bolukbasi, Liu, Xiong, Tenney, Andreas, and Guu]{akyurek2022towards}
Aky{\"u}rek, E., Bolukbasi, T., Liu, F., Xiong, B., Tenney, I., Andreas, J., and Guu, K.
\newblock Towards tracing knowledge in language models back to the training data.
\newblock In \emph{Findings of the Association for Computational Linguistics: EMNLP 2022}, pp.\  2429--2446. Association for Computational Linguistics, 2022.
\newblock \doi{10.18653/v1/2022.findings-emnlp.180}.
\newblock URL \url{https://aclanthology.org/2022.findings-emnlp.180/}.

\bibitem[Anshumann et~al.(2025)Anshumann, Zaidi, Kedia, Ahn, Kwon, Lee, Lee, and Lee]{anshumann2025sparselogitsamplingaccelerating}
Anshumann, Zaidi, M.~A., Kedia, A., Ahn, J., Kwon, T., Lee, K., Lee, H., and Lee, J.
\newblock Sparse logit sampling: Accelerating knowledge distillation in llms, 2025.
\newblock URL \url{https://arxiv.org/abs/2503.16870}.

\bibitem[Bercovich et~al.(2025)Bercovich, Levy, Golan, Dabbah, El-Yaniv, Puny, Galil, Moshe, Ronen, Nabwani, et~al.]{bercovich2025llama}
Bercovich, A., Levy, I., Golan, I., Dabbah, M., El-Yaniv, R., Puny, O., Galil, I., Moshe, Z., Ronen, T., Nabwani, N., et~al.
\newblock Llama-nemotron: Efficient reasoning models.
\newblock \emph{arXiv preprint arXiv:2505.00949}, 2025.

\bibitem[Carlini et~al.(2021)Carlini, Tramer, Wallace, Jagielski, Herbert-Voss, Lee, Roberts, Brown, Song, Erlingsson, et~al.]{carlini2021extracting}
Carlini, N., Tramer, F., Wallace, E., Jagielski, M., Herbert-Voss, A., Lee, K., Roberts, A., Brown, T., Song, D., Erlingsson, U., et~al.
\newblock Extracting training data from large language models.
\newblock In \emph{30th USENIX security symposium (USENIX Security 21)}, pp.\  2633--2650, 2021.

\bibitem[Carlini et~al.(2023)Carlini, Ippolito, Jagielski, Lee, Tram{\`e}r, and Zhang]{carlini2023quantifying}
Carlini, N., Ippolito, D., Jagielski, M., Lee, K., Tram{\`e}r, F., and Zhang, C.
\newblock Quantifying memorization across neural language models.
\newblock In \emph{International Conference on Learning Representations (ICLR)}, 2023.
\newblock URL \url{https://openreview.net/forum?id=TatRHT_1cK}.

\bibitem[Chang et~al.(2022)Chang, Yang, and Lee]{chang2022distilhubert}
Chang, H.-J., Yang, S.-w., and Lee, H.-y.
\newblock Distilhubert: Speech representation learning by layer-wise distillation of hidden-unit bert.
\newblock In \emph{ICASSP 2022-2022 IEEE International Conference on Acoustics, Speech and Signal Processing (ICASSP)}, pp.\  7087--7091. IEEE, 2022.

\bibitem[Chen et~al.(2025)Chen, Zhang, Hong, Kundu, and Wang]{chen2025sealsteerablereasoningcalibration}
Chen, R., Zhang, Z., Hong, J., Kundu, S., and Wang, Z.
\newblock Seal: Steerable reasoning calibration of large language models for free, 2025.
\newblock URL \url{https://arxiv.org/abs/2504.07986}.

\bibitem[Dai et~al.(2024)Dai, Deng, Zhao, Xu, Gao, Chen, Li, Zeng, Yu, Wu, Xie, Li, Huang, Luo, Ruan, Sui, and Liang]{dai2024deepseekmoeultimateexpertspecialization}
Dai, D., Deng, C., Zhao, C., Xu, R.~X., Gao, H., Chen, D., Li, J., Zeng, W., Yu, X., Wu, Y., Xie, Z., Li, Y.~K., Huang, P., Luo, F., Ruan, C., Sui, Z., and Liang, W.
\newblock Deepseekmoe: Towards ultimate expert specialization in mixture-of-experts language models, 2024.
\newblock URL \url{https://arxiv.org/abs/2401.06066}.

\bibitem[DeepSeek-AI(2025)]{deepseekai2025deepseekv3technicalreport}
DeepSeek-AI.
\newblock Deepseek-v3 technical report, 2025.
\newblock URL \url{https://arxiv.org/abs/2412.19437}.

\bibitem[Fedus et~al.(2022)Fedus, Zoph, and Shazeer]{fedus2022switchtransformersscalingtrillion}
Fedus, W., Zoph, B., and Shazeer, N.
\newblock Switch transformers: Scaling to trillion parameter models with simple and efficient sparsity, 2022.
\newblock URL \url{https://arxiv.org/abs/2101.03961}.

\bibitem[Gou et~al.(2021)Gou, Yu, Maybank, and Tao]{gou2021knowledge}
Gou, J., Yu, B., Maybank, S.~J., and Tao, D.
\newblock Knowledge distillation: A survey.
\newblock \emph{International journal of computer vision}, 129\penalty0 (6):\penalty0 1789--1819, 2021.

\bibitem[Guo et~al.(2025)Guo, Yang, Zhang, Song, Zhang, Xu, Zhu, Ma, Wang, Bi, et~al.]{guo2025deepseek}
Guo, D., Yang, D., Zhang, H., Song, J., Zhang, R., Xu, R., Zhu, Q., Ma, S., Wang, P., Bi, X., et~al.
\newblock Deepseek-r1: Incentivizing reasoning capability in llms via reinforcement learning.
\newblock \emph{arXiv preprint arXiv:2501.12948}, 2025.

\bibitem[Hendrycks et~al.(2021)Hendrycks, Basart, Kadavath, Mazeika, Arora, Guo, Burns, Puranik, He, Song, and Steinhardt]{hendrycksapps2021}
Hendrycks, D., Basart, S., Kadavath, S., Mazeika, M., Arora, A., Guo, E., Burns, C., Puranik, S., He, H., Song, D., and Steinhardt, J.
\newblock Measuring coding challenge competence with apps.
\newblock \emph{NeurIPS}, 2021.

\bibitem[Hinton et~al.(2015)Hinton, Vinyals, and Dean]{hinton2015distillingknowledgeneuralnetwork}
Hinton, G., Vinyals, O., and Dean, J.
\newblock Distilling the knowledge in a neural network, 2015.
\newblock URL \url{https://arxiv.org/abs/1503.02531}.

\bibitem[Hsieh et~al.(2023)Hsieh, Li, Yeh, Nakhost, Fujii, Ratner, Krishna, Lee, and Pfister]{hsieh2023distilling}
Hsieh, C.-Y., Li, C.-L., Yeh, C.-K., Nakhost, H., Fujii, Y., Ratner, A., Krishna, R., Lee, C.-Y., and Pfister, T.
\newblock Distilling step-by-step! outperforming larger language models with less training data and smaller model sizes.
\newblock \emph{arXiv preprint arXiv:2305.02301}, 2023.

\bibitem[Jiang et~al.(2024)Jiang, Sablayrolles, Roux, Mensch, Savary, Bamford, Chaplot, de~las Casas, Hanna, Bressand, Lengyel, Bour, Lample, Lavaud, Saulnier, Lachaux, Stock, Subramanian, Yang, Antoniak, Scao, Gervet, Lavril, Wang, Lacroix, and Sayed]{jiang2024mixtralexperts}
Jiang, A.~Q., Sablayrolles, A., Roux, A., Mensch, A., Savary, B., Bamford, C., Chaplot, D.~S., de~las Casas, D., Hanna, E.~B., Bressand, F., Lengyel, G., Bour, G., Lample, G., Lavaud, L.~R., Saulnier, L., Lachaux, M.-A., Stock, P., Subramanian, S., Yang, S., Antoniak, S., Scao, T.~L., Gervet, T., Lavril, T., Wang, T., Lacroix, T., and Sayed, W.~E.
\newblock Mixtral of experts, 2024.
\newblock URL \url{https://arxiv.org/abs/2401.04088}.

\bibitem[Kandpal et~al.(2022)Kandpal, Wallace, and Raffel]{kandpal2022deduplicating}
Kandpal, N., Wallace, E., and Raffel, C.
\newblock Deduplicating training data mitigates privacy risks in language models.
\newblock In \emph{Proceedings of the 39th International Conference on Machine Learning (ICML)}, volume 162 of \emph{Proceedings of Machine Learning Research}, pp.\  11220--11234. PMLR, 2022.
\newblock URL \url{https://proceedings.mlr.press/v162/kandpal22a/kandpal22a.pdf}.

\bibitem[Krishna et~al.(2019)Krishna, Tomar, Parikh, Papernot, and Iyyer]{krishna2019thieves}
Krishna, K., Tomar, G.~S., Parikh, A.~P., Papernot, N., and Iyyer, M.
\newblock Thieves on sesame street! model extraction of bert-based apis.
\newblock \emph{arXiv preprint arXiv:1910.12366}, 2019.

\bibitem[Lambert et~al.(2024)Lambert, Morrison, Pyatkin, Huang, Ivison, Brahman, Miranda, Liu, Dziri, Lyu, Gu, Malik, Graf, Hwang, Yang, Bras, Tafjord, Wilhelm, Soldaini, Smith, Wang, Dasigi, and Hajishirzi]{lambert2024tulu3}
Lambert, N., Morrison, J., Pyatkin, V., Huang, S., Ivison, H., Brahman, F., Miranda, L. J.~V., Liu, A., Dziri, N., Lyu, S., Gu, Y., Malik, S., Graf, V., Hwang, J.~D., Yang, J., Bras, R.~L., Tafjord, O., Wilhelm, C., Soldaini, L., Smith, N.~A., Wang, Y., Dasigi, P., and Hajishirzi, H.
\newblock Tülu 3: Pushing frontiers in open language model post-training.
\newblock 2024.

\bibitem[Lee et~al.(2022)Lee, Ippolito, Nystrom, Zhang, Eck, Callison-Burch, and Carlini]{lee2022deduplicating}
Lee, K., Ippolito, D., Nystrom, A., Zhang, C., Eck, D., Callison-Burch, C., and Carlini, N.
\newblock Deduplicating training data makes language models better.
\newblock In \emph{Proceedings of the 60th Annual Meeting of the Association for Computational Linguistics (Volume 1: Long Papers)}, pp.\  8424--8445. Association for Computational Linguistics, 2022.
\newblock \doi{10.18653/v1/2022.acl-long.577}.
\newblock URL \url{https://aclanthology.org/2022.acl-long.577/}.

\bibitem[Lee et~al.(2025)Lee, Zhou, Ao, Li, Du, He, Wu, Liu, Liu, Alinejad-Rokny, Yang, Liang, Wen, and Ni]{lee2025quantificationlargelanguagemodel}
Lee, S., Zhou, J., Ao, C., Li, K., Du, X., He, S., Wu, H., Liu, T., Liu, J., Alinejad-Rokny, H., Yang, M., Liang, Y., Wen, Z., and Ni, S.
\newblock Quantification of large language model distillation, 2025.
\newblock URL \url{https://arxiv.org/abs/2501.12619}.

\bibitem[Li et~al.(2023{\natexlab{a}})Li, Hammoud, Itani, Khizbullin, and Ghanem]{li2023camel}
Li, G., Hammoud, H. A. A.~K., Itani, H., Khizbullin, D., and Ghanem, B.
\newblock Camel: Communicative agents for "mind" exploration of large scale language model society, 2023{\natexlab{a}}.

\bibitem[LI et~al.(2024)LI, Beeching, Tunstall, Lipkin, Soletskyi, Huang, Rasul, Yu, Jiang, Shen, Qin, Dong, Zhou, Fleureau, Lample, and Polu]{numina_math_datasets}
LI, J., Beeching, E., Tunstall, L., Lipkin, B., Soletskyi, R., Huang, S.~C., Rasul, K., Yu, L., Jiang, A., Shen, Z., Qin, Z., Dong, B., Zhou, L., Fleureau, Y., Lample, G., and Polu, S.
\newblock Numinamath.
\newblock \url{[https://huggingface.co/AI-MO/NuminaMath-CoT](https://github.com/project-numina/aimo-progress-prize/blob/main/report/numina_dataset.pdf)}, 2024.

\bibitem[Li et~al.(2023{\natexlab{b}})Li, Zhang, Yadav, Sung, Cheng, Bansal, and Chen]{li2023merge}
Li, P., Zhang, Z., Yadav, P., Sung, Y.-L., Cheng, Y., Bansal, M., and Chen, T.
\newblock Merge, then compress: Demystify efficient smoe with hints from its routing policy.
\newblock \emph{arXiv preprint arXiv:2310.01334}, 2023{\natexlab{b}}.

\bibitem[Li et~al.(2024)Li, Zhang, Yadav, Sung, Cheng, Bansal, and Chen]{li2024mergecompressdemystifyefficient}
Li, P., Zhang, Z., Yadav, P., Sung, Y.-L., Cheng, Y., Bansal, M., and Chen, T.
\newblock Merge, then compress: Demystify efficient smoe with hints from its routing policy, 2024.
\newblock URL \url{https://arxiv.org/abs/2310.01334}.

\bibitem[Li et~al.(2025{\natexlab{a}})Li, Jin, Tan, Cheng, and Chen]{li2025quantmoebenchexaminingposttrainingquantization}
Li, P., Jin, X., Tan, Z., Cheng, Y., and Chen, T.
\newblock Quantmoe-bench: Examining post-training quantization for mixture-of-experts, 2025{\natexlab{a}}.
\newblock URL \url{https://arxiv.org/abs/2406.08155}.

\bibitem[Li et~al.(2025{\natexlab{b}})Li, Tan, Qu, Liu, and Chen]{li2025doge}
Li, P., Tan, Z., Qu, H., Liu, H., and Chen, T.
\newblock Doge: Defensive output generation for llm protection against knowledge distillation.
\newblock \emph{arXiv preprint arXiv:2505.19504}, 2025{\natexlab{b}}.

\bibitem[Li et~al.(2023{\natexlab{c}})Li, Fu, Zhang, Huang, Sun, Lyu, Liu, Jin, and Li]{li2023taco}
Li, R., Fu, J., Zhang, B.-W., Huang, T., Sun, Z., Lyu, C., Liu, G., Jin, Z., and Li, G.
\newblock Taco: Topics in algorithmic code generation dataset.
\newblock \emph{arXiv preprint arXiv:2312.14852}, 2023{\natexlab{c}}.

\bibitem[Li et~al.(2022)Li, Choi, Chung, Kushman, Schrittwieser, Leblond, Eccles, Keeling, Gimeno, Dal~Lago, Hubert, Choy, de~Masson~d'Autume, Babuschkin, Chen, Huang, Welbl, Gowal, Cherepanov, Molloy, Mankowitz, Sutherland~Robson, Kohli, de~Freitas, Kavukcuoglu, and Vinyals]{li2022competition}
Li, Y., Choi, D., Chung, J., Kushman, N., Schrittwieser, J., Leblond, R., Eccles, T., Keeling, J., Gimeno, F., Dal~Lago, A., Hubert, T., Choy, P., de~Masson~d'Autume, C., Babuschkin, I., Chen, X., Huang, P.-S., Welbl, J., Gowal, S., Cherepanov, A., Molloy, J., Mankowitz, D., Sutherland~Robson, E., Kohli, P., de~Freitas, N., Kavukcuoglu, K., and Vinyals, O.
\newblock Competition-level code generation with alphacode.
\newblock \emph{arXiv preprint arXiv:2203.07814}, 2022.

\bibitem[Liang et~al.(2023)Liang, Zuo, Zhang, He, Chen, and Zhao]{liang2023less}
Liang, C., Zuo, S., Zhang, Q., He, P., Chen, W., and Zhao, T.
\newblock Less is more: Task-aware layer-wise distillation for language model compression.
\newblock In \emph{International Conference on Machine Learning}, pp.\  20852--20867. PMLR, 2023.

\bibitem[Lin et~al.(2021)Lin, Wu, Yang, Lee, and Ren]{lin-etal-2021-riddlesense}
Lin, B.~Y., Wu, Z., Yang, Y., Lee, D.-H., and Ren, X.
\newblock Riddlesense: Reasoning about riddle questions featuring linguistic creativity and commonsense knowledge.
\newblock 2021.

\bibitem[Lin et~al.(2023)Lin, Chen, and Kao]{lin2023lad}
Lin, Y.-J., Chen, K.-Y., and Kao, H.-Y.
\newblock Lad: Layer-wise adaptive distillation for bert model compression.
\newblock \emph{Sensors}, 23\penalty0 (3):\penalty0 1483, 2023.

\bibitem[Liu et~al.(2025)Liu, Su, Yao, Jiang, Lai, Du, Qin, Xu, Lu, Yan, Chen, Zheng, Liu, Liu, Yin, He, Zhu, Wang, Wang, Dong, Zhang, Kang, Zhang, Xu, Zhang, Wu, Zhou, and Yang]{liu2025muonscalablellmtraining}
Liu, J., Su, J., Yao, X., Jiang, Z., Lai, G., Du, Y., Qin, Y., Xu, W., Lu, E., Yan, J., Chen, Y., Zheng, H., Liu, Y., Liu, S., Yin, B., He, W., Zhu, H., Wang, Y., Wang, J., Dong, M., Zhang, Z., Kang, Y., Zhang, H., Xu, X., Zhang, Y., Wu, Y., Zhou, X., and Yang, Z.
\newblock Muon is scalable for llm training, 2025.
\newblock URL \url{https://arxiv.org/abs/2502.16982}.

\bibitem[Luo et~al.(2025{\natexlab{a}})Luo, Li, Peng, Wang, Cheng, Chen, et~al.]{luo2025occult}
Luo, S., Li, P., Peng, J., Wang, H., Cheng, Y., Chen, T., et~al.
\newblock Occult: Optimizing collaborative communication across experts for accelerated parallel moe training and inference.
\newblock \emph{arXiv preprint arXiv:2505.13345}, 2025{\natexlab{a}}.

\bibitem[Luo et~al.(2025{\natexlab{b}})Luo, Li, Peng, Zhao, Cao, Cheng, and Chen]{anonymous2025occult}
Luo, S., Li, P., Peng, J., Zhao, Y., Cao, Y., Cheng, Y., and Chen, T.
\newblock Occult: Optimizing collaborative communications across experts for accelerated parallel moe training and inference.
\newblock In \emph{Forty-second International Conference on Machine Learning}, 2025{\natexlab{b}}.
\newblock URL \url{https://openreview.net/forum?id=vh2Dt4sT67}.

\bibitem[Ma et~al.(2021)Ma, Chen, Hu, You, Xie, and Wang]{ma2021undistillablemakingnastyteacher}
Ma, H., Chen, T., Hu, T.-K., You, C., Xie, X., and Wang, Z.
\newblock Undistillable: Making a nasty teacher that cannot teach students, 2021.
\newblock URL \url{https://arxiv.org/abs/2105.07381}.

\bibitem[Ma et~al.(2022)Ma, Huang, Chen, Tang, You, Wang, and Xie]{ma2022stingy}
Ma, H., Huang, Y., Chen, T., Tang, H., You, C., Wang, Z., and Xie, X.
\newblock Stingy teacher: Sparse logits suffice to fail knowledge distillation, 2022.
\newblock URL \url{https://openreview.net/forum?id=ae7BJIOxkxH}.

\bibitem[Maini et~al.(2021)Maini, Yaghini, and Papernot]{maini2021dataset}
Maini, P., Yaghini, M., and Papernot, N.
\newblock Dataset inference: Ownership resolution in machine learning.
\newblock \emph{arXiv preprint arXiv:2104.10706}, 2021.

\bibitem[Mattern et~al.(2023)Mattern, Mireshghallah, Jin, Sch{\"o}lkopf, Sachan, and Berg-Kirkpatrick]{mattern2023membership}
Mattern, J., Mireshghallah, F., Jin, Z., Sch{\"o}lkopf, B., Sachan, M., and Berg-Kirkpatrick, T.
\newblock Membership inference attacks against language models via neighbourhood comparison.
\newblock \emph{arXiv preprint arXiv:2305.18462}, 2023.

\bibitem[Muennighoff et~al.(2024)Muennighoff, Soldaini, Groeneveld, Lo, Morrison, Min, Shi, Walsh, Tafjord, Lambert, Gu, Arora, Bhagia, Schwenk, Wadden, Wettig, Hui, Dettmers, Kiela, Farhadi, Smith, Koh, Singh, and Hajishirzi]{muennighoff2024olmoeopenmixtureofexpertslanguage}
Muennighoff, N., Soldaini, L., Groeneveld, D., Lo, K., Morrison, J., Min, S., Shi, W., Walsh, P., Tafjord, O., Lambert, N., Gu, Y., Arora, S., Bhagia, A., Schwenk, D., Wadden, D., Wettig, A., Hui, B., Dettmers, T., Kiela, D., Farhadi, A., Smith, N.~A., Koh, P.~W., Singh, A., and Hajishirzi, H.
\newblock Olmoe: Open mixture-of-experts language models, 2024.
\newblock URL \url{https://arxiv.org/abs/2409.02060}.

\bibitem[Muennighoff et~al.(2025)Muennighoff, Yang, Shi, Li, Fei-Fei, Hajishirzi, Zettlemoyer, Liang, Cand{\`e}s, and Hashimoto]{muennighoff2025s1}
Muennighoff, N., Yang, Z., Shi, W., Li, X.~L., Fei-Fei, L., Hajishirzi, H., Zettlemoyer, L., Liang, P., Cand{\`e}s, E., and Hashimoto, T.
\newblock s1: Simple test-time scaling.
\newblock \emph{arXiv preprint arXiv:2501.19393}, 2025.

\bibitem[Park et~al.(2023)Park, Georgiev, Ilyas, Leclerc, and Madry]{park2023trak}
Park, S.~M., Georgiev, K., Ilyas, A., Leclerc, G., and Madry, A.
\newblock {TRAK}: Attributing model behavior at scale.
\newblock In \emph{Proceedings of the 40th International Conference on Machine Learning (ICML)}, volume 202 of \emph{Proceedings of Machine Learning Research}. PMLR, 2023.
\newblock URL \url{https://proceedings.mlr.press/v202/park23c/park23c.pdf}.

\bibitem[Penedo et~al.(2025)Penedo, Lozhkov, Kydlíček, Allal, Beeching, Lajarín, Gallouédec, Habib, Tunstall, and von Werra]{penedo2025codeforces}
Penedo, G., Lozhkov, A., Kydlíček, H., Allal, L.~B., Beeching, E., Lajarín, A.~P., Gallouédec, Q., Habib, N., Tunstall, L., and von Werra, L.
\newblock Codeforces.
\newblock \url{https://huggingface.co/datasets/open-r1/codeforces}, 2025.

\bibitem[Pruthi et~al.(2020)Pruthi, Liu, Kale, and Sundararajan]{pruthi2020tracin}
Pruthi, G., Liu, F., Kale, S., and Sundararajan, M.
\newblock Estimating training data influence by tracing gradient descent.
\newblock In \emph{Advances in Neural Information Processing Systems (NeurIPS)}, 2020.
\newblock URL \url{https://proceedings.neurips.cc/paper/2020/file/e6385d39ec9394f2f3a354d9d2b88eec-Paper.pdf}.

\bibitem[Qiu et~al.(2025)Qiu, Guo, Song, Sun, Cai, Wei, Luo, Yin, Zhang, Hu, et~al.]{qiu2025phybench}
Qiu, S., Guo, S., Song, Z.-Y., Sun, Y., Cai, Z., Wei, J., Luo, T., Yin, Y., Zhang, H., Hu, Y., et~al.
\newblock Phybench: Holistic evaluation of physical perception and reasoning in large language models.
\newblock \emph{arXiv preprint arXiv:2504.16074}, 2025.

\bibitem[Sanh et~al.(2019)Sanh, Debut, Chaumond, and Wolf]{sanh2019distilbert}
Sanh, V., Debut, L., Chaumond, J., and Wolf, T.
\newblock Distilbert, a distilled version of bert: smaller, faster, cheaper and lighter.
\newblock \emph{arXiv preprint arXiv:1910.01108}, 2019.

\bibitem[Savani et~al.(2025)Savani, Trockman, Feng, Schwarzschild, Robey, Finzi, and Kolter]{savani2025antidistillationsampling}
Savani, Y., Trockman, A., Feng, Z., Schwarzschild, A., Robey, A., Finzi, M., and Kolter, J.~Z.
\newblock Antidistillation sampling, 2025.
\newblock URL \url{https://arxiv.org/abs/2504.13146}.

\bibitem[Shazeer et~al.(2017)Shazeer, Mirhoseini, Maziarz, Davis, Le, Hinton, and Dean]{shazeer2017outrageouslylargeneuralnetworks}
Shazeer, N., Mirhoseini, A., Maziarz, K., Davis, A., Le, Q., Hinton, G., and Dean, J.
\newblock Outrageously large neural networks: The sparsely-gated mixture-of-experts layer, 2017.
\newblock URL \url{https://arxiv.org/abs/1701.06538}.

\bibitem[Sun et~al.(2025)Sun, Yin, Xu, Kolter, and Liu]{sun2025idiosyncrasieslargelanguagemodels}
Sun, M., Yin, Y., Xu, Z., Kolter, J.~Z., and Liu, Z.
\newblock Idiosyncrasies in large language models, 2025.
\newblock URL \url{https://arxiv.org/abs/2502.12150}.

\bibitem[Wang \& Yoon(2021)Wang and Yoon]{wang2021knowledge}
Wang, L. and Yoon, K.-J.
\newblock Knowledge distillation and student-teacher learning for visual intelligence: A review and new outlooks.
\newblock \emph{IEEE transactions on pattern analysis and machine intelligence}, 44\penalty0 (6):\penalty0 3048--3068, 2021.

\bibitem[Warner et~al.(2024)Warner, Chaffin, Clavié, Weller, Hallström, Taghadouini, Gallagher, Biswas, Ladhak, Aarsen, Cooper, Adams, Howard, and Poli]{modernbert}
Warner, B., Chaffin, A., Clavié, B., Weller, O., Hallström, O., Taghadouini, S., Gallagher, A., Biswas, R., Ladhak, F., Aarsen, T., Cooper, N., Adams, G., Howard, J., and Poli, I.
\newblock Smarter, better, faster, longer: A modern bidirectional encoder for fast, memory efficient, and long context finetuning and inference, 2024.
\newblock URL \url{https://arxiv.org/abs/2412.13663}.

\bibitem[Wei et~al.(2024)Wei, Zhu, Zhao, Cheng, Li, Lü, Cheng, Zhang, Zhang, Zeng, Wang, Ma, Hu, Yan, Fang, and Zhou]{wei2024skyworkmoedeepdivetraining}
Wei, T., Zhu, B., Zhao, L., Cheng, C., Li, B., Lü, W., Cheng, P., Zhang, J., Zhang, X., Zeng, L., Wang, X., Ma, Y., Hu, R., Yan, S., Fang, H., and Zhou, Y.
\newblock Skywork-moe: A deep dive into training techniques for mixture-of-experts language models, 2024.
\newblock URL \url{https://arxiv.org/abs/2406.06563}.

\bibitem[Xu et~al.(2024)Xu, Li, Tao, Shen, Cheng, Li, Xu, Tao, and Zhou]{xu2024survey}
Xu, X., Li, M., Tao, C., Shen, T., Cheng, R., Li, J., Xu, C., Tao, D., and Zhou, T.
\newblock A survey on knowledge distillation of large language models.
\newblock \emph{arXiv preprint arXiv:2402.13116}, 2024.

\bibitem[Yang et~al.(2025)Yang, Li, Yang, Zhang, Hui, Zheng, Yu, Gao, Huang, Lv, et~al.]{yang2025qwen3}
Yang, A., Li, A., Yang, B., Zhang, B., Hui, B., Zheng, B., Yu, B., Gao, C., Huang, C., Lv, C., et~al.
\newblock Qwen3 technical report.
\newblock \emph{arXiv preprint arXiv:2505.09388}, 2025.

\bibitem[Yang et~al.(2024)Yang, Zhu, Lu, Wang, Chen, Gao, Yan, and Chen]{yang2024survey}
Yang, C., Zhu, Y., Lu, W., Wang, Y., Chen, Q., Gao, C., Yan, B., and Chen, Y.
\newblock Survey on knowledge distillation for large language models: methods, evaluation, and application.
\newblock \emph{ACM Transactions on Intelligent Systems and Technology}, 2024.

\bibitem[Yu et~al.(2024)Yu, Lin, Yu, and Xing]{yu2024gptfuzzerredteaminglarge}
Yu, J., Lin, X., Yu, Z., and Xing, X.
\newblock Gptfuzzer: Red teaming large language models with auto-generated jailbreak prompts, 2024.
\newblock URL \url{https://arxiv.org/abs/2309.10253}.

\bibitem[Zhang et~al.(2025)Zhang, Li, Peng, Qiu, and Chen]{zhang2025advancingmoeefficiencycollaborationconstrained}
Zhang, M., Li, P., Peng, J., Qiu, M., and Chen, T.
\newblock Advancing moe efficiency: A collaboration-constrained routing (c2r) strategy for better expert parallelism design, 2025.
\newblock URL \url{https://arxiv.org/abs/2504.01337}.

\end{thebibliography}
\bibliographystyle{style/icml2025}

\titlespacing*{\section}{0pt}{*1}{*1}
\titlespacing*{\subsection}{0pt}{*1.25}{*1.25}
\titlespacing*{\subsubsection}{0pt}{*1.5}{*1.5}

\setlength{\abovedisplayskip}{\baselineskip} 
\setlength{\abovedisplayshortskip}{0.5\baselineskip} 
\setlength{\belowdisplayskip}{\baselineskip}
\setlength{\belowdisplayshortskip}{0.5\baselineskip}

\clearpage
\appendix
\label{sec:append}
\part*{Appendix}
{
\setlength{\parskip}{-0em}
\startcontents[sections]
\printcontents[sections]{ }{1}{}
}

\setlength{\parskip}{.5em}
\section{Experiment Details}

Experiments were conducted on NVIDIA A100 and B200 GPU servers. For all training runs, we use the AdamW optimizer with a weight decay of $0.01$ and a warm-up ratio of $0.1$. For all MoE models, we apply a load-balancing loss with a coefficient of $0.001$. We apply all distillation experiments for $3$ epochs with the learning rate of $5 \times 10^{-6}$ and the batch size of $256$. We apply cosine learning rate schedulers. 

\section{Dataset Details}

We list the datasets we used in this work and their license here:
\begin{itemize}
    \item Tulu3~\citep{lambert2024tulu3} with ODC-BY-1.0 license.
    \item TACO~\citep{li2023taco} with Apache 2.0 license.
    \item Apps~\citep{hendrycksapps2021} with MIT license.
    \item Code Contests~\citep{li2022competition} with CC-by-4.0 license
    \item Codeforces~\citep{penedo2025codeforces} with CC-by-4.0 license
    \item NuminaMath~\citep{numina_math_datasets} with Apache 2.0 license
    \item Chemistry~\citep{li2023camel} with CC-by-NC-4.0 license
    \item Biology~\citep{li2023camel} with CC-by-NC-4.0 license
    \item Physics~\citep{li2023camel} with CC-by-NC-4.0 license
    \item Riddle Sense~\citep{lin-etal-2021-riddlesense}
\end{itemize}

\section{Details of Distance Metrics for Routing Pattern Comparison}
\label{sec:distance_metrics}

In this section, we provide detailed mathematical formulations and computational procedures for the Wasserstein distance metrics used to compare expert routing patterns between models.

\subsection{Expert Specialization Distance}

Given two models (teacher $g_T$ and student $g_S$) with expert specialization profiles $\tilde{S}^{(\ell)}_T$ and $\tilde{S}^{(\ell)}_S$ at layer $\ell$ (\Cref{def:special}), we compute the permutation-invariant Wasserstein distance to measure their similarity. For a specific domain $d \in \mathcal{D}$, the normalized specialization profiles $\tilde{S}^{(\ell)}_T[:, d] \in \Delta^{E_\ell}$ and $\tilde{S}^{(\ell)}_S[:, d] \in \Delta^{E_\ell}$ represent probability distributions over $E_\ell$ experts, where each column sums to 1 as specified in \Cref{def:special}.

The Wasserstein-1 distance between two discrete distributions on expert indices is computed as:
\begin{equation}
W_1(\tilde{S}^{(\ell)}_T[:, d], \tilde{S}^{(\ell)}_S[:, d]) = \inf_{\gamma \in \Gamma} \sum_{i=1}^{E_\ell} \sum_{j=1}^{E_\ell} |i - j| \cdot \gamma_{i,j}
\end{equation}
where $\Gamma = \Gamma(\tilde{S}^{(\ell)}_T[:, d], \tilde{S}^{(\ell)}_S[:, d])$ is the set of all joint distributions with marginals $\tilde{S}^{(\ell)}_T[:, d]$ and $\tilde{S}^{(\ell)}_S[:, d]$. In practice, we use the optimal transport formulation implemented in \texttt{scipy.stats.wasserstein\_distance}, which takes expert positions $\mathbf{pos} = [0, 1, \ldots, E_\ell - 1]$ as the ground metric.

Since expert indices are arbitrary permutations of the same underlying functionality, we compute the optimal permutation-invariant distance as defined in \Cref{eq:d_spec}:
\begin{equation}
d^{(\ell)}_{\text{spec}} = \min_{\Pi \in \Pi_{E_\ell}} \frac{1}{D} \sum_{d=1}^{D} W_1\left(\Pi \tilde{S}^{(\ell)}_T[:, d], \tilde{S}^{(\ell)}_S[:, d]\right)
\end{equation}
where $\Pi_{E_\ell}$ denotes the set of all $E_\ell \times E_\ell$ permutation matrices. The optimization over permutations is solved using the Hungarian algorithm, which finds the optimal assignment in $\mathcal{O}(E_\ell^3)$ time by minimizing the total cost across all domains.

\subsection{Expert Collaboration Distance}

For expert collaboration patterns $\tilde{B}^{(\ell)}_T$ and $\tilde{B}^{(\ell)}_S$ (\Cref{def:collab}), we measure similarity through permutation-invariant Wasserstein distance. The normalized collaboration matrix $\tilde{B}^{(\ell)} \in [0, 1]^{E_\ell \times E_\ell}$ captures pairwise expert co-activation frequencies, where $\sum_{i \neq j} \tilde{B}^{(\ell)}_{i,j} = 1$ and the diagonal is zero. Each row $\tilde{B}^{(\ell)}[i, :]$ represents the probability distribution of expert $i$ collaborating with other experts.

To compute the Wasserstein distance between collaboration patterns, we treat the collaboration matrix as a collection of probability distributions. For computational efficiency, we represent the collaboration patterns as sparse dictionaries mapping expert pairs to co-occurrence probabilities:
\begin{equation}
\mathcal{B}_T = \{(i, j) \mapsto \tilde{B}^{(\ell)}_{T, i,j} : i \neq j, i,j \in [E_\ell]\}
\end{equation}

For a specific expert $i$, we extract the row vector $\tilde{B}^{(\ell)}_T[i, :]$ and compute its Wasserstein distance to the corresponding row in the student model. The computation proceeds as follows. First, identify all expert pairs that have non-zero collaboration in either model:
\begin{equation}
\mathcal{P}_i = \{j : \tilde{B}^{(\ell)}_{T, i,j} > 0 \text{ or } \tilde{B}^{(\ell)}_{S, i,j} > 0, j \neq i\}
\end{equation}
Second, construct aligned probability vectors by extracting collaboration probabilities for all pairs in $\mathcal{P}_i$, with missing entries defaulting to zero, then normalize to ensure valid probability distributions:
\begin{equation}
\mathbf{p}_{T,i} = [\tilde{B}^{(\ell)}_{T, i,j_1}, \ldots, \tilde{B}^{(\ell)}_{T, i,j_{|\mathcal{P}_i|}}]^T, \quad \hat{\mathbf{p}}_{T,i} = \frac{\mathbf{p}_{T,i}}{\|\mathbf{p}_{T,i}\|_1}
\end{equation}
with analogous construction for $\hat{\mathbf{p}}_{S,i}$. Third, compute the Wasserstein distance using position indices:
\begin{equation}
W_1(\tilde{B}^{(\ell)}_T[i, :], \tilde{B}^{(\ell)}_S[i, :]) = \text{wasserstein\_distance}([0, \ldots, |\mathcal{P}_i|-1], [0, \ldots, |\mathcal{P}_i|-1], \hat{\mathbf{p}}_{T,i}, \hat{\mathbf{p}}_{S,i})
\end{equation}

Following \Cref{eq:d_collab}, the permutation-invariant collaboration distance averages over all expert rows after applying the optimal permutation:
\begin{equation}
d^{(\ell)}_{\text{collab}} = \min_{\Pi \in \Pi_{E_\ell}} \frac{1}{E_\ell} \sum_{i=1}^{E_\ell} W_1\left((\Pi \tilde{B}^{(\ell)}_T \Pi^T)[i, :], \tilde{B}^{(\ell)}_S[i, :]\right)
\end{equation}
where $\Pi \tilde{B}^{(\ell)}_T \Pi^T$ applies the same permutation to both rows and columns of the collaboration matrix to maintain consistency in expert indexing.

\subsection{Aggregate Detection Score}

The final detection score combines both specialization and collaboration distances from the last MoE layer $\ell = L$:
\begin{equation}
s(f_T, f_S) = -\frac{1}{2}\left(d^{(L)}_{\text{spec}} + d^{(L)}_{\text{collab}}\right)
\end{equation}
where higher scores indicate stronger routing similarity and thus higher likelihood of a distillation relationship. In the pairwise classification task (\Cref{sec:special-collab}), given a candidate pair $\{f^{\text{KD}}_S, f^{\text{scratch}}_S\}$, we select the model with the higher score as the distilled candidate:
\begin{equation}
\hat{\imath} = \argmax_{i \in \{\text{KD}, \text{scratch}\}} s(f_T, f^{(i)}_S)
\end{equation}

The computational complexity is $\mathcal{O}(E_L^3 \cdot D)$ for specialization (Hungarian algorithm over $D$ domains) and $\mathcal{O}(E_L^4)$ for collaboration (permutation matching over $E_L$ expert rows), yielding a total complexity of $\mathcal{O}(E_L^4 + E_L^3 D)$ per model pair. For our experiments with $E_L = 64$ experts and $D = 9$ domains, the computation completes within minutes on standard hardware.

\end{document}